\documentclass[acmsmall,screen]{acmart}
\AtBeginDocument{%
  }

\setcopyright{acmlicensed}
\copyrightyear{2025}
\acmYear{2025}
\acmDOI{XXXXXXX.XXXXXXX}

\acmJournal{JACM}
\acmVolume{XX}
\acmNumber{XX}
\acmArticle{XX}
\acmMonth{1}

\usepackage{hyperref}
\usepackage{url}
\usepackage{xcolor}         

\usepackage[ruled,vlined]{algorithm2e}
\usepackage{subcaption}
\usepackage{threeparttable}
\usepackage{multirow}
\usepackage{multicol}
\usepackage{listings}
\usepackage{enumitem}

\definecolor{keywordcolor}{rgb}{0.5,0,0.5} 
\definecolor{stringcolor}{rgb}{0.25,0.5,0.35} 
\definecolor{commentcolor}{rgb}{0.5,0.5,0.5} 
\definecolor{bgcolor}{rgb}{0.95,0.95,0.92} 
\definecolor{numbercolor}{rgb}{0.7,0.7,0.7} 
\newcommand{\evorl}{\texttt{\textbf{EvoRL}}}

\definecolor{16n16e}{rgb}{0.12156862745098039, 0.4666666666666667, 0.7058823529411765}
\definecolor{16n1e}{rgb}{1.0, 0.4980392156862745, 0.054901960784313725}
\definecolor{256n16e}{rgb}{0.17254901960784313, 0.6274509803921569, 0.17254901960784313}
\definecolor{256n1e}{rgb}{0.8392156862745098, 0.15294117647058825, 0.1568627450980392}

\begin{document}

\title{EvoRL: A GPU-accelerated Framework for Evolutionary Reinforcement Learning}


\author{Bowen Zheng}
\email{bowen.zheng@protonmail.com}
\affiliation{%
  \department{Department of Data Science and Artificial Intelligence}
  \institution{The Hong Kong Polytechnic University}
  \city{Hong Kong SAR}
  \country{China}
}

\author{Ran Cheng}
\authornote{Corresponding Author}
\email{ranchengcn@gmail.com}
\affiliation{%
  \department{Department of Data Science and Artificial Intelligence, and Department of Computing}
  \institution{The Hong Kong Polytechnic University}
  \city{Hong Kong SAR}
  \country{China}
}
\affiliation{
  \institution{The Hong Kong Polytechnic University Shenzhen Research Institute}
  \city{Shenzhen}
  \country{China}
}

\author{Kay Chen Tan}
\email{kctan@polyu.edu.hk}
\affiliation{%
  \department{Department of Data Science and Artificial Intelligence}
  \institution{The Hong Kong Polytechnic University}
  \city{Hong Kong SAR}
  \country{China}
}

\renewcommand{\shortauthors}{Zheng \textit{et al.}}

\begin{abstract}
    Evolutionary Reinforcement Learning (EvoRL) has emerged as a promising approach to overcoming the limitations of traditional reinforcement learning (RL) by integrating the Evolutionary Computation (EC) paradigm with RL. However, the population-based nature of EC significantly increases computational costs, thereby restricting the exploration of algorithmic design choices and scalability in large-scale settings.
    To address this challenge, we introduce \evorl{}\footnote{To distinguish the framework from the general term EvoRL, we use \evorl{} to refer specifically to the proposed framework.}, the first end-to-end EvoRL framework optimized for GPU acceleration. The framework executes the entire training pipeline on accelerators, including environment simulations and EC processes, leveraging hierarchical parallelism through vectorization and compilation techniques to achieve superior speed and scalability. This design enables the efficient training of large populations on a single machine.
    In addition to its performance-oriented design, \evorl{} offers a comprehensive platform for EvoRL research, encompassing implementations of traditional RL algorithms (e.g., A2C, PPO, DDPG, TD3, SAC), Evolutionary Algorithms (e.g., CMA-ES, OpenES, ARS), and hybrid EvoRL paradigms such as Evolutionary-guided RL (e.g., ERL, CEM-RL) and Population-Based AutoRL (e.g., PBT). 
    The framework's modular architecture and user-friendly interface allow researchers to seamlessly integrate new components, customize algorithms, and conduct fair benchmarking and ablation studies. 
    The project is open-source and available at: \url{https://github.com/EMI-Group/evorl}.
\end{abstract}

\ccsdesc[500]{Computing methodologies~Reinforcement learning}
\ccsdesc[500]{Computing methodologies~Evolutionary robotics}

\keywords{Evolutionary Computation, Evolutionary Algorithm, Evolutionary Reinforcement Learning}


\maketitle

\section{Introduction}
\label{sec:intro}

Reinforcement Learning (RL) is a machine learning paradigm aimed at training agents for sequential decision-making tasks. Unlike supervised learning, which operates on static datasets, RL learns an agent's policy through interactions with the environment, seeking to maximize cumulative rewards (returns). Advances in gradient-based optimization within deep learning have driven RL to achieve significant success across various domains, including board games \citep{silverMasteringGameGo2016,silverGeneralReinforcementLearning2018}, video games \citep{mnihHumanlevelControlDeep2015,mnihAsynchronousMethodsDeep2016,vinyalsGrandmasterLevelStarCraft2019}, and robotic control \citep{lillicrapContinuousControlDeep2016,fujimotoAddressingFunctionApproximation2018,haarnojaSoftActorCriticAlgorithms2019}. 
Despite its successes, RL faces critical challenges due to the non-stationary nature of its training data source. These challenges include instability arising from sensitivity to hyperparameters and the need to carefully balance exploration and exploitation. 
Beyond RL, the Evolutionary Computation (EC) has emerged as a competitive alternative for policy search. EC leverages population-based methods, e.g., Evolutionary Algorithms (EAs), which utilize episodic returns as feedback signals and ensure diverse exploration through the population. However, EAs are often constrained by high sample complexity and the inefficiencies of their gradient-free, heuristic-driven optimization strategies, resulting in slow convergence rates.

Recent research has focused on integrating EC and RL methodologies, leveraging their unique strengths to address individual limitations. 
A pioneering work in this area is Evolution-guided Reinforcement Learning (ERL) \citep{khadkaEvolutionguidedPolicyGradient2018}, which combines EC and off-policy RL using a shared replay buffer. 
In this framework, the RL policy benefits from the diverse exploration facilitated by the population of an EA, whose trajectories are stored in the replay buffer. 
Conversely, the population is guided by the RL policy to enhance overall performance. 
Building upon ERL, several variants \citep{pourchotCEMRLCombiningEvolutionary2019,haoERLRe2EfficientEvolutionary2022,liValueEvolutionaryBasedReinforcementLearning2024} have been proposed, demonstrating promising results across a variety of tasks.
Beyond this paradigm, due to the black-box characteristics, EAs can also be employed to optimize different RL components in an automated reinforcement learning (AutoRL) framework \citep{parker-holderAutomatedReinforcementLearning2022}. 
For instance, Population-Based Training (PBT) \citep{jaderbergPopulationBasedTraining2017} uses a population-based approach for dynamic hyperparameter tuning, training a group of agents with different hyperparameters and evolving these hyperparameters during the training process. 
This approach effectively stabilizes training and enhances final performance.
In summary, these methods leverage the diversity of populations and the gradient-free optimization properties of EAs, resulting in powerful hybrid frameworks. 
Collectively, we refer to these methods as Evolutionary Reinforcement Learning (EvoRL).

Although EvoRL holds significant potential, several limitations continue to hinder its further development. 
Compared to traditional RL methods, the computational cost of EvoRL is notably higher due to the additional sample complexity and optimization costs introduced by population-based approaches. 
Previous studies, particularly within the ERL paradigm, have focused on designing sample-efficient hybrid algorithms. 
However, this focus often results in constrained population sizes and the introduction of additional mechanisms, which come at the cost of increased training time \citep{bodnarProximalDistilledEvolutionary2020,haoERLRe2EfficientEvolutionary2022}. 
Consequently, large-scale settings, such as those involving substantial population sizes, remain largely unexplored.
Even under constrained population sizes, the training costs associated with EvoRL remain significant. 
As a result, many design choices and mechanisms in EvoRL algorithms have not been thoroughly investigated. For instance, most studies have only examined a limited set of EAs, and additional mechanisms and corresponding hyperparameters governing the integration of information between EC and RL require further analysis.

In summary, the limitations can be attributed to the following factors: (1) The environments are primarily executed on CPUs, while the agents' decision-making and learning processes occur on heterogeneous devices such as GPUs. This setup necessitates frequent and dense communication between CPUs and GPUs, consuming a significant portion of training time. (2) CPU-based environments lack the scalability required to efficiently support EvoRL under large population sizes. (3) Additionally, operators in EC are often executed on CPUs or inefficiently implemented, further exacerbating computational bottlenecks. (4) Other components in the training pipeline also suffer from inefficiencies; for instance, the population's training process is frequently executed through sequential orders, such as \emph{for loops} in Python, which fails to fully exploit the parallel computing capabilities of GPUs. While asynchronous techniques can potentially improve efficiency, they significantly increase the complexity of writing and debugging EvoRL algorithms.

\begin{figure}
    \centering
     \includegraphics[width=\linewidth]{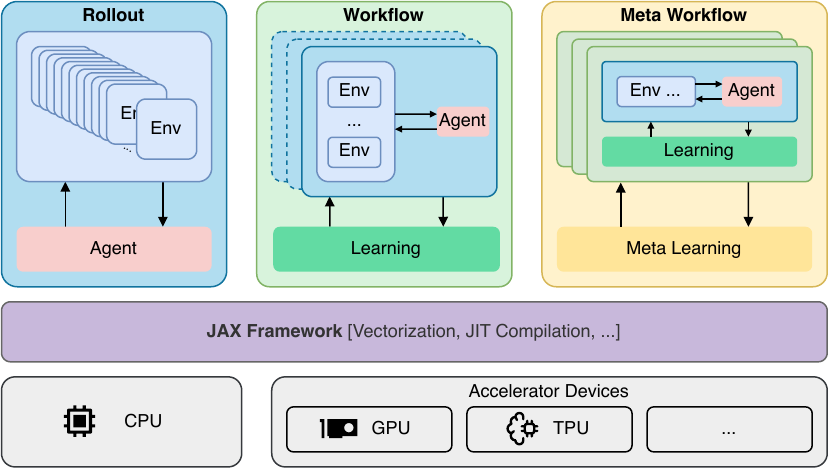}
    \caption{The hierarchical vectorization architecture of \evorl{}. Three levels of vectorization are listed from left to right: parallel environments, parallel agents, and parallel training. It efficiently executes procedures in EvoRL algorithms in parallel, fully utilizing the computing capability of modern accelerator architectures.}
    \label{fig:hierarchy-vectorization}
\end{figure}

To address these challenges and advance future research in EvoRL, we propose a new framework named \evorl{}. 
Designed for high efficiency, scalability, and ease of use, \evorl{} is the first end-to-end EvoRL framework that fully integrates EC with RL. 
The framework relocates the entire training pipeline on GPUs, including the execution of environments and EAs.
To maximize computational efficiency, \evorl{} employs vectorization techniques to enable hierarchical parallelism across three dimensions: parallel environments, parallel agents, and parallel training, as shown in Fig.~\ref{fig:hierarchy-vectorization}.
This hierarchical vectorization architecture significantly improves scalability and computational performance, meeting the demands of EvoRL algorithms while fully utilizing the parallel computing capabilities of modern GPU architectures. 
Additionally, compilation techniques are seamlessly integrated throughout the training pipeline to further enhance performance and efficiency.
As a result, EvoRL algorithms can be executed efficiently even on a single GPU while maintaining support for large population sizes. 
Complementing these technical advancements, \evorl{} provides a user-friendly interface, simplifying the integration of EC and RL and enabling researchers to effortlessly design, train, and evaluate EvoRL algorithms.

Currently, \evorl{} supports canonical RL and EC for policy search and offers two widely adopted EvoRL paradigms: ERL and Population-Based AutoRL (see Sec.~\ref{sec:bg-evorl}). The framework provides a set of modular interfaces, enabling users to easily replace and customize different components, such as the EC components in ERL. 
For Population-Based AutoRL, we have implemented PBT for hyperparameter tuning and generalized its explore-and-exploit procedure.
This includes a customizable evolutionary layer to support various selection and mutation operators within the EC paradigm. 
As an open-source framework, \evorl{} aims to lower the barriers to developing efficient EvoRL algorithms while serving as a foundational platform for future research.
With the rapid emergence of GPU-accelerated environments and the promising potential of EvoRL, we believe that this framework is both timely and necessary.
Our contributions are summarized as follows:
\begin{enumerate}[nolistsep,leftmargin=*]
    \item {\evorl{} provides the first end-to-end EvoRL framework optimized for GPU acceleration.} It seamlessly integrates EC, RL, and simulated environments on GPUs,  fully leveraging the parallelism of modern hardware and eliminating the overhead caused by CPU-based communication.
    
    \item {\evorl{} employs advanced vectorization and compilation techniques.} These techniques are applied throughout the training pipeline, enabling significant speed-ups and scalability to support large-scale training settings, such as substantial population sizes. While maintaining high efficiency, \evorl{} provides a user-friendly platform for researchers to design and develop EvoRL algorithms.
    
    \item {\evorl{} offers a diverse suite of baseline algorithms.} In addition to canonical EAs and RL algorithms, \evorl{} covers end-to-end training pipelines for ERL and its variants. Furthermore, it provides a general PBT workflow that is compatible with other pipelines, accompanied by examples for both on-policy and off-policy RL algorithms.
\end{enumerate}

\section{Background}
\label{sec:bg}

\subsection{Reinforcement Learning}

Model-free reinforcement learning (RL) is generally framed under the Markov Decision Process (MDP) assumption \citep{suttonReinforcementLearningIntroduction2018}, where the problem is defined by a tuple $(\mathcal{S}, \mathcal{A}, p, r)$. $\mathcal{S}$ is the state space, $\mathcal{A}$ is the action space, $p: \mathcal{S} \times \mathcal{S} \times \mathcal{A} \mapsto [0,1]$ is the transition probability function, and $r: \mathcal{S} \times \mathcal{A} \mapsto [r_{\text{min}},r_{\text{max}}]$ is the reward function after every transition. The goal is to find an optimal policy $\pi(a_t|s_t)$ that maximizes the expected return: $\mathbb{E}_{\tau \sim p(\tau|\pi)}  \sum_{t}^{|\tau|}  r_t $, where $\tau$ is a trajectory generated by the policy $\pi$. In addition, a discount factor $\gamma \in [0,1]$ is introduced as an inductive bias\footnote{Although, in the episodic tasks, the discount factor has been treated as a part of the environment previously, it is deemed as a hyperparameter of RL algorithms \citep{hesselInductiveBiasesDeep2019,amitDiscountFactorRegularizer2020,grigsbyAutomaticActorCriticSolutions2021} in recent research, where agents' undiscounted returns are reported for evaluation.}.


Unlike supervised learning with static data sources, RL algorithms need to balance between exploitation and exploration.
Exploitation is efficiently addressed through gradient-based updates when paired with delicate implementation and hyperparameter tuning. Remarkable successes have been achieved across multiple domains~\citep{mnihHumanlevelControlDeep2015,vinyalsGrandmasterLevelStarCraft2019,fujimotoAddressingFunctionApproximation2018,ganesh2019reinforcement,aradi2022survey}. However, the "deadly triad,"~\citep{suttonReinforcementLearningIntroduction2018,hasselt2018deep} a combination of function approximation, bootstrapping, and off-policy learning, can cause instability or even failure during temporal difference learning. On the other hand, exploration strategies, which decide how agents explore unknown scenarios, remain an open topic. Several additional mechanisms \citep{ostrovski2017countbased,osband2016deep,houthooft2016vime,raffin2021smooth} have been proposed to mitigate the issue of exploration diversity.

\subsection{Evolutionary Algorithms for Policy Search}

Due to their gradient-free and black-box characteristics, Evolutionary Computation (EC) is highly versatile and finds broad applicability across a wide range of domains. 
As EC methods, the Evolutionary Algorithms (EAs) are popular alternatives for policy search and have demonstrated performance comparable to RL across various tasks \citep{salimansEvolutionStrategiesScalable2017,maniaSimpleRandomSearch2018,chenRestartbasedRank1Evolution2019}.

EAs perform direct policy optimization based on the episodic returns of agents within a population and inherently utilize population-based exploration in the parameter space. 
Unlike RL methods, EAs have a built-in mechanism for balancing exploitation and exploration, which helps reduce training noise to stabilize optimization. 
Furthermore, EAs do not rely on the temporal structure of episodic trajectories, enabling them to discard additional components required by RL, such as the Markov Decision Process (MDP) assumption and the discount factor.
This flexibility allows EAs to efficiently handle delayed or sparse rewards without the need for high-quality reward shaping functions in RL algorithms.
Additionally, with the incorporation of quality diversity techniques \citep{pughQualityDiversityNew2016}, population-based exploration in EAs can address deceptive rewards and mitigate the risk of convergence to local optima.

However, these advantages come with notable limitations. EAs often suffer from high sample complexity, particularly concerning the number of episodic trajectories for fitness evaluation \citep{salimansEvolutionStrategiesScalable2017,chrabaszczBackBasicsBenchmarking2018}. Moreover, as gradient-free optimization methods, EAs typically yield lower performance compared to state-of-the-art gradient-based RL methods.
Despite these limitations, the finding of \citet{salimansEvolutionStrategiesScalable2017} suggests the intrinsic dimensionality of RL problems may be significantly smaller than expected, challenging previous assumptions regarding the curse of dimensionality in EAs. It provides the feasibility of using EAs to optimize neural networks for policy search.

\subsection{Evolutionary Reinforcement Learning}
\label{sec:bg-evorl}

Given the complementary characteristics of EC and RL, recent research has focused on hybrid approaches that combine the strengths of both paradigms while mitigating their respective limitations. 
These methods, collectively referred to as Evolutionary Reinforcement Learning (EvoRL), extend the flexibility of EC to enhance the RL training pipeline from multiple perspectives.

In EvoRL, EC and RL can be used simultaneously to optimize policy parameters, mutually assisting each other. This paradigm is referred to as Evolution-guided Reinforcement Learning (ERL). For instance, the original ERL framework \citep{khadkaEvolutionguidedPolicyGradient2018} incorporates a single RL agent alongside a population of agents managed by an EA, both of which are trained in parallel and share a replay buffer to store transitions. The RL agent is trained using data from the shared replay buffer, while the population is optimized based on guidance from the RL policy. Subsequent variants enhance the assistance between EC and RL. For example, CEM-RL \citep{pourchotCEMRLCombiningEvolutionary2019} leverages RL as a local search operator to update half of the population in each iteration, while Supe-RL \citep{marchesini2021genetic} directly transfers elite policy parameters from an EA to the RL agent. ERL-RL$^2$ \citep{haoERLRe2EfficientEvolutionary2022} takes a different approach by sharing a nonlinear state representation between policy networks optimized by EC and RL, while optimizing independent linear policy layers. Additionally, RL can serve as an efficient operator to accelerate quality diversity algorithms, as demonstrated in \citet{nilssonPolicyGradientAssisted2021}.

Another prominent EvoRL paradigm employs EAs as meta-algorithms for RL, referred to as population-based AutoRL. A notable example is Population-Based Training (PBT) \citep{jaderbergPopulationBasedTraining2017}, which proposes an online hyperparameter tuning method. PBT trains a population of models with varying hyperparameters, where each individual consists of both hyperparameters and network weights. After each training interval, PBT selects high-performing individuals to replace poorly performing ones based on the meta objectives of each individual, such as the average evaluation return. Then it perturbs the hyperparameters of the replaced individuals. PBT can be viewed as a specific instance of EC \citep{petrenkoDexPBTScalingDexterous2023,shahidScalingPopulationBasedReinforcement2024}, incorporating mutation and evolutionary selection operators. For example, \citet{baiGeneralizedPopulationBasedTraining2024} introduced a pairwise learning operator inspired by the Cooperative Swarm Optimizer (CSO) \citep{chengCompetitiveSwarmOptimizer2015} to replace the standard mutation operator in PBT. 
Similarly, \citet{dushatskiyMultiObjectivePopulationBased2023} extended PBT into multi-objective scenarios using the non-dominated ranking approach from NSGA-II \citep{debFastElitistMultiobjective2002}. 
Furthermore, EAs have been used as black-box optimizers to approximate meta-gradients, which are often intractable in RL scenarios \citep{tangOnlineHyperparameterTuning2020}, offering a promising direction for hyperparameter tuning in RL.

Beyond hyperparameter tuning, EC has also been employed to search for reward functions \citep{faustEvolvingRewardsAutomate2019,saporaEvILEvolutionStrategies2024,onori2024adaptive} or loss functions \citep{houthooftEvolvedPolicyGradients2018,luDiscoveredPolicyOptimisation2022} in RL algorithms. Other integration strategies include using EC as a mechanism for RL action selection \citep{maEvolutionaryActionSelection2022}. For a more comprehensive review of these approaches, we refer readers to recent surveys \citep{baiEvolutionaryReinforcementLearning2023,linEvolutionaryReinforcementLearning2024,liBridgingEvolutionaryAlgorithms2024,sigaudCombiningEvolutionDeep2023}.

\subsection{GPU-Accelerated Environments}
\label{sec:gpu-envs}

Simulators play a pivotal role in RL research. 
On the one hand, some environments, such as video games and board games, naturally operate on simulators. 
On the other hand, simulators provide approximations for studying real-world tasks, such as robotic control. 
Traditional simulators, including Mujoco \citep{todorovMuJoCoPhysicsEngine2012}, PyBullet \citep{coumans2021}, and DART \citep{leeDARTDynamicAnimation2018}, are primarily implemented on CPUs.\footnote{Technically, Mujoco 3.0 includes the MuJoCo XLA (MJX) module accelerated by JAX, and PyBullet offers experimental OpenCL GPGPU support.} 
However, in RL training pipelines, the agents' decisions and learning processes typically occur on heterogeneous devices like GPUs, creating a bottleneck due to the frequent and dense communication between CPUs and GPUs.
Moreover, these simulators consume substantial CPU resources, particularly in large-scale training settings \citep{horganDistributedPrioritizedExperience2018,salimansEvolutionStrategiesScalable2017,chrabaszczBackBasicsBenchmarking2018}.

A viable solution to eliminate CPU-GPU bottlenecks in RL training is to implement simulations directly on GPUs, which are naturally suited for high-throughput parallel simulations. \citet{liangGPUAcceleratedRoboticSimulation2018} firstly demonstrated the promising speed-up achieved by GPU-accelerated robotic simulations in RL. Expanding on this concept, Isaac Gym \citep{makoviychukIsaacGymHigh2021} introduced an end-to-end GPU-based RL training pipeline, effectively eliminating the communication overhead between CPUs and GPUs. Compared to traditional CPU-based simulators, Isaac Gym achieves a 100x–1000x speed-up in training. Several subsequent studies \citep{petrenkoDexPBTScalingDexterous2023,handa2023dextreme,shahidScalingPopulationBasedReinforcement2024} have further validated the effectiveness of this end-to-end training pipeline using Isaac Gym. More recently, Genesis \citep{Genesis} introduced even more efficient and universal physics engines with GPU support, further advancing the development of GPU-accelerated RL environments. Besides, Mujoco playground \citep{mujoco_playground_2025} has recently been released for GPU-accelerated environments for a comprehensive study in robot learning.

In recent years, the adoption of RL ecosystems based on JAX has surged. JAX is a high-performance scientific computing library that runs on accelerators such as GPUs and TPUs, offering features like automatic differentiation and vectorization. 
This enables the entire RL training pipeline to be natively implemented in JAX for improved efficiency. Several GPU-accelerated environments have emerged within this ecosystem. 
For instance, Brax \citep{freemanBraxDifferentiablePhysics2021} provides a 3D physics simulator for continuous control tasks, serving as a replacement for the original Mujoco. 
Jax2D \citep{matthewsKinetixInvestigatingTraining2024} extends the functionality to 2D physics simulations. 
Beyond physics simulations, JAX-based environments encompass a diverse range of applications, including video games \citep{gymnax2022github,rutherfordJaxMARLMultiAgentRL2024}, board games \citep{koyamadaPgxHardwareAcceleratedParallel2023}, maze navigation \citep{jiangMinimaxEfficientBaselines2024}, and combinatorial optimization problems \citep{bonnetJumanjiDiverseSuite2024}.

The primary speed-up achieved by GPU-accelerated environments stems from their ability to run a massive number of environment instances in parallel. 
This contradicts traditional RL algorithms, which typically utilize only a few parallel environments. 
As a result, special algorithmic designs are necessary to efficiently handle the large-scale input data generated by these environments and achieve fast convergence. 
While several algorithms, such as IMPALA \citep{espeholtIMPALAScalableDistributed2018} and Ape-X \citep{horganDistributedPrioritizedExperience2018}, have been proposed to address this challenge, the problem remains an open area for future research. 
Nonetheless, we believe that integrating EC into RL presents a promising direction to tackle this challenge.

\subsection{GPU-Accelerated Evolutionary Algorithms}

Traditionally, EAs have been implemented on CPUs, with notable libraries such as PlatEMO \citep{tianPlatEMOMATLABPlatform2017}, Pymoo \citep{blankPymooMultiObjectiveOptimization2020}, and DEAP \citep{fortinDEAPEvolutionaryAlgorithms2012}. 
However, the remarkable success of gradient-based deep learning, powered by GPUs, has inspired the development of GPU-accelerated EAs. 
By leveraging the massive parallel compute units of GPUs, EAs can significantly improve their efficiency, particularly when scaling up the search space or increasing population size.

GPU acceleration benefits several components of EAs, such as random noise generation and matrix operations, which achieve substantial speed-ups compared to traditional CPU-based implementations. 
Furthermore, GPU-based EAs can seamlessly integrate gradient-based operations, enabling hybrid approaches that combine evolutionary search with gradient-based optimization \citep{gangwani2018policy}.
Additionally, GPU-accelerated EAs benefit from problems implemented on GPUs, facilitating the development of end-to-end training pipelines that eliminate the overhead caused by frequent communication between CPUs and GPUs.

To support these advancements, several libraries have been developed. 
EvoTorch \citep{tokluEvoTorchScalableEvolutionary2023} utilizes PyTorch to build scalable EAs, while libraries such as EvoJax \citep{tangEvoJAXHardwareAcceleratedNeuroevolution2022}, evosax \citep{langeEvosaxJAXbasedEvolution2022}, and EvoX \citep{huangEvoXDistributedGPUaccelerated2023} implement GPU-accelerated EAs in JAX. 
These libraries not only provide high-performance implementations but also enable seamless integration with existing machine learning frameworks, further advancing the capabilities of EAs in large-scale optimization tasks.

\section{Architecture}

The design of \evorl{} aims to address the challenges of implementing scalable and efficient EvoRL algorithms by providing a modular, extensible, and high-performance framework. 
Built on Python with JAX, \evorl{} is specifically designed to leverage the computational advantages of accelerators like GPUs while maintaining ease of use for researchers. 
The architecture adopts a functional programming approach, emphasizing modularity and composability to support a wide range of RL, EC, and hybrid algorithmic designs.
With an emphasis on seamless integration with JAX-based ecosystems and advanced parallelism techniques, \evorl{} enables the development of complex EvoRL training pipelines with minimal overhead. 
This section outlines the core components of \evorl{}, illustrating how its architecture supports flexibility, scalability, and efficiency in the EvoRL research workflow.

\subsection{Programming Model}

\begin{figure}
    \centering
    \includegraphics[width=\linewidth]{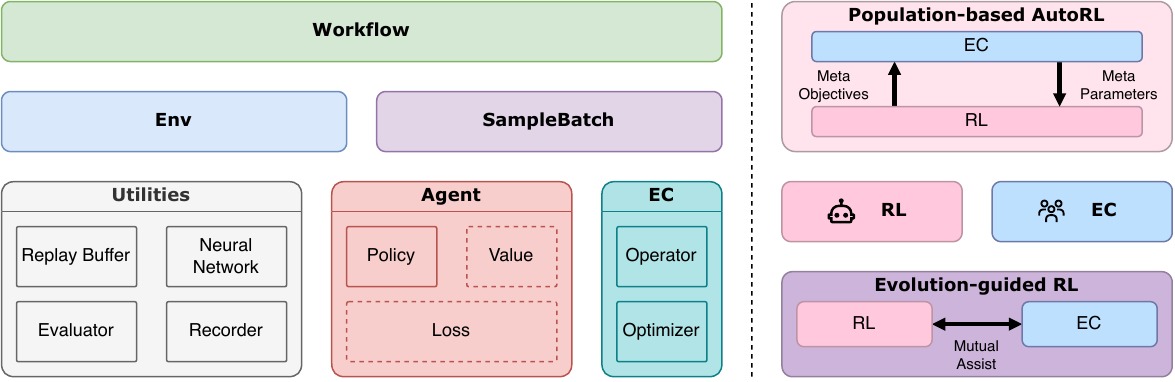}
    \caption{\textbf{Left}: Core components of \evorl{}. \textbf{Right}: Supported algorithm types in \evorl{} for policy search. Currently, \evorl{} includes Reinforcement Learning (RL): PPO, DQN, TD3, \textit{et~ al.}, Evolutionary Computation (EC): OpenES, ARS, CMA-ES, \textit{et~ al.}, Evolution-guided Reinforcement Learning (ERL): original ERL, CEM-RL, and related variants, Population-based AutoRL: PBT and its variants for hyperparameter tuning.}
    \label{fig:evorl-components}
\end{figure}

\evorl{} is designed to simplify the development of EvoRL algorithms and their associated components. It provides comprehensive support for building RL algorithms, EAs, and hybrid EvoRL algorithms. 
Implemented in Python with JAX, \evorl{} integrates seamlessly with the JAX-based RL ecosystem, encouraging collaboration within its active community. 
The framework employs an object-oriented functional programming model, where classes define the execution logic while maintaining the state externally. This design enhances modularity and composability, with its key components illustrated in Fig.~\ref{fig:evorl-components} and summarized as follows:

\begin{itemize}[nolistsep,leftmargin=*]
    \item \lstinline|Env|: This class provides a unified interface to interact with various JAX-based environments from multiple libraries, ensuring consistency and ease of integration. It abstracts away the specifics of the underlying environment implementations, allowing seamless integration into the training pipelines. In addition, general-purpose environmental wrappers are provided. For instance, the \lstinline|VmapAutoReset| wrapper runs multiple environment copies and
    automatically resets them upon termination, ensuring uninterrupted data collection during training.
    
    \item \lstinline|Agent|: This class encapsulates the learning agent and defines its behavior for both training and evaluation. It manages key components, including the policy network, which determines the agent's decisions for actions, and an optional value network used for estimating state or state-action values. The class also specifies optional loss functions required for gradient-based updates in some algorithms. 
    
    \item \lstinline|SampleBatch|: This class is a flexible data structure designed to store and manage transitions generated from interactions between the agent and the environment. It can contain continuous trajectories or shuffled transition batches, catering to diverse algorithmic needs.

    \item \lstinline|Workflow|: This class defines the overarching training logic for algorithms implemented in \evorl{}. Each algorithm is built by deriving from the \lstinline|Workflow| class, where the \lstinline|step()| method encapsulates a single training iteration. This method is designed to leverage JAX's Just-In-Time (JIT) compilation and vectorization capabilities, ensuring efficient execution on modern accelerators like GPUs. The \lstinline|learn()| method orchestrates the entire training loop, managing key tasks such as termination condition checks, performance evaluation, periodic logging, and model checkpointing.
    
    \item \lstinline|EC|: This module provides a comprehensive suite of components for EC, including various EAs and related operators such as mutation and selection. These EAs can be directly used as policy optimizers in EC pipelines. Along with the operators, they can be smoothly integrated into the RL workflows to construct or modify complex EvoRL algorithms.
    
    \item \lstinline|Utilities|: A rich collection of utilities is provided to streamline the implementation of training pipelines, including:
        \begin{itemize}[nolistsep,leftmargin=*]
            \item A GPU VRAM-based replay buffer designed for end-to-end pipelines to avoid frequent CPU-GPU communication.
            \item Neural network toolkits that simplify the construction of efficient and flexible model architectures. These toolkits are optimized for EvoRL scenarios and integrate seamlessly with the framework.
            \item Versatile logging tools, which allow users to monitor various metrics during training, such as rewards, losses, and hyperparameter updates. These logs are customizable and can be outputted into different formats for analysis and visualization.
            \item The \lstinline|Evaluator| module, which is designed to assess agent performance comprehensively. It includes specialized variants capable of collecting trajectories from a population of agents to satisfy the demand for some hybrid algorithms like ERL.
        \end{itemize}
\end{itemize}

Examples of training pipelines implemented using the \lstinline|Workflow| class are depicted in Fig.~\ref{fig:train-logic}. Additional code examples are available in Appendix~\ref{sec:example-codes}.

\begin{figure}
    \centering
    \begin{subfigure}[c]{0.478\linewidth}
        \centering
        \includegraphics[width=\linewidth]{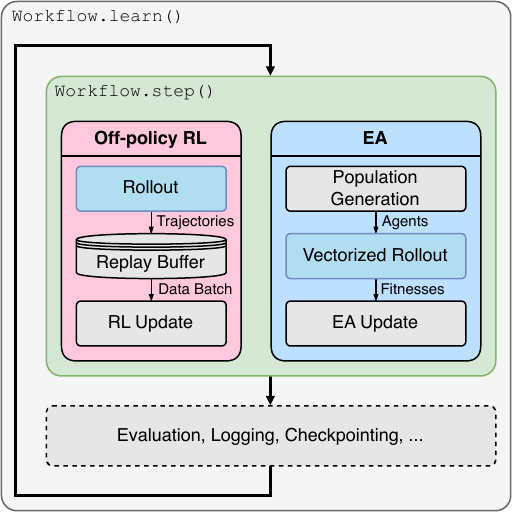}
        \caption{}
        \label{fig:basic-train-logic}
    \end{subfigure}
    \hfill
    \begin{subfigure}[c]{0.25\linewidth}
        \centering
        \includegraphics[width=\linewidth]{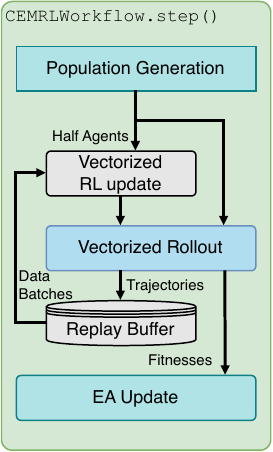}
        \caption{}
        \label{fig:cemrl-train-logic}
    \end{subfigure}
    \hfill
    \begin{subfigure}[c]{0.25\linewidth}
        \centering
        \includegraphics[width=\linewidth]{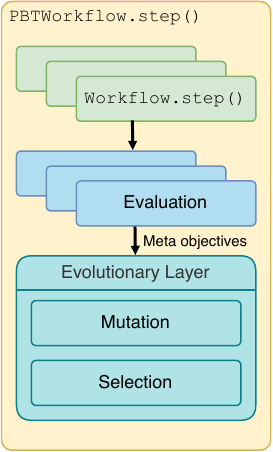}
        \caption{}
        \label{fig:pbt-train-logic}
    \end{subfigure}
    \caption{\textbf{(a)}: Example training logics in \evorl{}. Each algorithm has its own \lstinline|Workflow| class with customizable \lstinline|step()| for its training logic. The procedures of one step in the off-policy RL algorithm and the evolutionary algorithm are shown. \textbf{(b)}: A complex ERL workflow example. The training logic of CEM-RL is demonstrated.
    \textbf{(c)}: A population-based AutoRL workflow example for PBT. Meta Workflows are derived from \lstinline|Workflow|, sharing the same interface.}
    \label{fig:train-logic}
\end{figure}

\subsection{Computation Model}

The architecture of \evorl{} is designed under the assumption that the computational cost of a single environment and the scale of neural networks used in agents remain relatively small, as is typical in prior studies. 
Unlike supervised learning, where larger neural networks often lead to better performance, RL does not inherently benefit from large network architectures without additional considerations \citep{otaFrameworkTrainingLarger2024}. 
As a result, the memory and computational requirements in traditional RL training pipelines often fail to fully utilize the parallel computing capabilities of modern accelerators like GPUs.

Vectorization has emerged as a promising technique to address these limitations in RL~\citep{flajoletFastPopulationBasedReinforcement2022}. 
While vectorization increases linear memory consumption compared to sequential execution, it is highly effective for policy search tasks involving smaller neural networks. 
In \evorl{}, we employ \lstinline|jax.vmap()| to implement vectorization, which enables automatic batching of transformations within complex functions. 
This approach fully leverages modern accelerator features, including high-bandwidth memory (e.g., GDDR or HBM), parallel processing architectures (e.g., SIMD, SIMT), and specialized matrix computation units such as Tensor Cores in Nvidia GPUs.

To accommodate the diverse requirements of different algorithms, \evorl{} introduces a hierarchical vectorization architecture, illustrated in Fig.~\ref{fig:hierarchy-vectorization}. This architecture operates across three levels of parallelism:
\begin{enumerate}[nolistsep,leftmargin=*]
    \item \textbf{Parallel Environments}: Neural networks within agents inherently maintain a batch dimension, so vectorization focuses on the environments to enable batched observations and rewards during rollouts (i.e., interactions between agents and environments).
    \item \textbf{Parallel Agents}: EAs and ERL algorithms often involve evaluations of multiple independent agents. 
    To support this, we vectorize the rollout process across multiple agents, thus allowing simultaneous interaction with environments.
    \item \textbf{Parallel Training}: In ERL algorithms, vectorization extends to parallelize the training process across multiple agents. 
    Similarly, for population-based AutoRL algorithms, the entire training logic, encapsulated in \lstinline|Workflow.step()|, is vectorized to execute in parallel.
\end{enumerate}


The hierarchical vectorization architecture plays a critical role in supporting efficient parallelism in various algorithmic components from different types of algorithms, as illustrated in Fig.~\ref{fig:train-logic}. For example, evolutionary algorithms (EAs) require simultaneous evaluation of all agents in the population to compute fitness scores, where each agent interacts with multiple parallel environment copies to collect episode returns. This necessitates a two-level vectorization scheme to manage both agent-level and environment-level parallelism. In the case of CEM-RL, an even more complex vectorization strategy is employed in multiple dimensions. A population of agents is initialized, and at each iteration, half agents are randomly selected and updated using off-policy reinforcement learning (RL). After this, all agents undergo an evaluation to perform an EA update, and their batched trajectories are stored in a replay buffer for subsequent off-policy RL updates. Population-Based Training (PBT) further extends the need for hierarchical vectorization, as it orchestrates multiple concurrent workflows. It also requires parallel evaluations to compute meta-objectives (e.g., the average episode return) for evolutionary tuning of hyperparameters.

Besides hierarchical vectorization, we also apply JIT compilation throughout the entire training logic in \lstinline|Workflow.step()|, which effectively optimizes the tiny or redundant operations in the computation graph of the training pipeline. 
By combining these two techniques, along with the modular design, \evorl{} achieves highly reusable components that enable the development of efficient, scalable, and readable algorithms. 
This design not only maximizes the computational potential of modern hardware but also simplifies the implementation of complex training pipelines, making \evorl{} an effective framework for diverse EvoRL research needs.

\section{Platform}

Building upon the tailored architecture, \evorl{} provides a comprehensive platform for the development, experimentation, and evaluation of EvoRL algorithms. 
The supported algorithm types are summarized in the right panel of Fig.~\ref{fig:evorl-components}, involving EC, RL, and EvoRL. 

For RL, \evorl{} supports a wide range of traditional algorithms, including on-policy methods such as A2C \citep{mnihAsynchronousMethodsDeep2016} and PPO \citep{schulman2017proximal}, as well as off-policy algorithms like DQN \citep{mnihHumanlevelControlDeep2015}, IMPALA \citep{espeholtIMPALAScalableDistributed2018}, DDPG \citep{lillicrapContinuousControlDeep2016}, TD3 \citep{fujimotoAddressingFunctionApproximation2018}, and SAC \citep{haarnojaSoftActorCriticAlgorithms2019}. These implementations cover both discrete and continuous action spaces, enabling \evorl{} to address diverse problem domains, from game-playing to robotic control tasks.

For EC, \evorl{} includes robust implementations of widely used EAs such as CMA-ES \citep{hansenCMAEvolutionStrategy2016}, OpenES \citep{salimansEvolutionStrategiesScalable2017}, VanillaES \citep{chrabaszczBackBasicsBenchmarking2018}, and ARS \citep{maniaSimpleRandomSearch2018}. 
Additionally, the framework provides an adapter for algorithms in EvoX \citep{huangEvoXDistributedGPUaccelerated2023}, ensuring seamless integration and compatibility with external evolutionary libraries.

For EvoRL, \evorl{} supports two primary paradigms: Evolutionary-guided RL (ERL) and Population-Based AutoRL. Supported ERL implementations include the original ERL algorithm \citep{khadkaEvolutionguidedPolicyGradient2018}, CEM-RL \citep{pourchotCEMRLCombiningEvolutionary2019} algorithm, and their variants. 
For Population-Based AutoRL, \evorl{} includes PBT \citep{jaderbergPopulationBasedTraining2017} and its extensions. 
These algorithms serve as strong baselines, and \evorl{}’s modular and extensible architecture ensures that additional algorithms can be seamlessly integrated, fostering rapid innovation and experimentation.

The importance of implementation in RL has been highlighted in previous work \citep{engstromImplementationMattersDeep2020,andrychowiczWhatMattersOnPolicy2020}. 
However, ERL research often relies on original codebases for comparative studies without careful alignment, leading to inconsistent results.
\evorl{} addresses this issue by providing a unified, fair, and standardized platform for benchmarking and ablation studies. 
Its modular architecture allows researchers to easily integrate new components into existing pipelines.
For example, the default Cross-Entropy Method (CEM) in CEM-RL (see Fig.~\ref{fig:cemrl-train-logic}) can be replaced with alternative evolutionary algorithms from the EC module without requiring significant modifications to the pipeline.
Additionally, \evorl{}’s efficiency accelerates the trial-and-error process in research, enabling rapid exploration of new ideas and the ability to scale to larger configurations, such as increased population sizes or extended training durations.

\section{Experiments}

In this section, we comprehensively evaluate the performance, scalability, and efficiency of various algorithms implemented in \evorl{}. Our experiments focus on continuous robotic locomotion tasks in Brax \citep{freemanBraxDifferentiablePhysics2021}, a GPU-accelerated physics engine. 
Since these tasks are good replacements for the original CPU-based
counterpart on Gym \citep{brockmanOpenAIGym2016}, they serve as a robust benchmark for assessing \evorl{}’s ability to leverage GPU acceleration and handle large-scale computational workloads, ensuring fair comparisons with other state-of-the-art implementations.

The primary objectives of our experiments are as follows:
\begin{enumerate}[nolistsep,leftmargin=*]
    \item \textbf{Evaluating computational efficiency:} We analyze the efficiency of \evorl{}’s fully GPU-accelerated pipeline across diverse algorithmic paradigms, including EAs, ERL algorithms, and Population-based AutoRL algorithms. These results are benchmarked against existing implementations such as RLlib, which utilizes distributed CPU-based pipelines or hybrid pipelines leveraging both CPUs and GPUs.
    
    \item \textbf{Benchmarking algorithms:} We reproduce and validate the performance of different types of algorithms implemented in \evorl{}. By aligning hyperparameter settings and evaluation protocols, we ensure reliable comparisons and establish strong baselines for future research in EvoRL.
    

\end{enumerate}

All experiments are conducted under controlled settings to ensure fairness and reproducibility. To provide a consistent basis for comparison, we adhere to standardized hyperparameters and evaluation criteria across all algorithms and implementations. The modular design of \evorl{} facilitates the integration of diverse algorithmic components, enabling direct alignment with baseline implementations and isolating the contributions of specific architectural enhancements.

\subsection{Acceleration and Scalability Performance}
\label{sec:speed-exp}

\evorl{} constructs an end-to-end training pipeline for various algorithms, optimized for execution on modern accelerators such as GPUs. To evaluate the efficiency of these pipelines, we measure the average training time across different configurations and compare the results to prior implementations. For consistency, we use the Swimmer environment from both Gym \citep{brockmanOpenAIGym2016} and Brax \citep{freemanBraxDifferentiablePhysics2021}, which features fixed-length episodes (1000 timesteps). This choice eliminates potential variability due to agent performance across implementations, ensuring fair pipeline-level comparisons. All experiments are conducted on a single machine equipped with dual Intel Xeon Gold 6132 CPUs (56 logical cores), 128 GiB RAM, and an Nvidia RTX 3090 GPU with 24 GiB VRAM. Detailed training settings and hyperparameters are provided in Appendix~\ref{sec:training-settings}. In addition, Table~\ref{tab:speed_benchmark_config} summarizes the essential training pipeline differences across different libraries.

\begin{table}[h]
    \centering
    \caption{Training pipeline implementations on different libraries.}
    \label{tab:speed_benchmark_config}
    \begin{subtable}[t]{\linewidth}
        \centering
        \begin{tabular}{|c|c|c|c|}
            \hline
            Library & Environment & Agent Update  & Rollout \\ \hline
            EvoRL & Brax Swimmer & GPU & GPU Vectorization \\ \hline
            RLlib & Gym Swimmer-v4 & CPU & Multiprocess (CPU) \\ \hline
        \end{tabular}
        \caption{EA Implementation}
    \end{subtable}
    \begin{subtable}[t]{\linewidth}
        \centering
        \begin{tabular}{|c|c|c|c|}
            \hline
            Library & Environment & Agent Update  & Rollout \\ \hline
            EvoRL & Brax Swimmer & GPU Vectorization & GPU Vectorization \\ \hline
            Official & Gym Swimmer-v4 & For-loop (GPU) & For-loop (Sinlge CPU) \\ \hline
            fastpbrl & Gym Swimmer-v4 & GPU Vectorization & Multiprocess (CPU) \\ \hline
        \end{tabular}
        \caption{CEM-RL Implementation}
    \end{subtable}
    \begin{subtable}[t]{\linewidth}
        \centering
        \begin{tabular}{|c|c|c|c|}
            \hline
            Library & Environment & Agent Update  & Rollout \\ \hline
            EvoRL & Brax Swimmer & GPU Vectorization & GPU Vectorization \\ \hline
            fastpbrl & Gym Swimmer-v4 & GPU Vectorization & Multiprocess (CPU) \\ \hline
            RLLib (PBT) & Gym Swimmer-v4 & GPU Multiprocess & Multiprocess (CPU) \\ \hline
        \end{tabular}
        \caption{PBT Implementation}
    \end{subtable}
\end{table}

\subsubsection{Performance of EAs for Policy Search}
\label{sec:speed-exp-ea}

\begin{figure}[t]
    \centering
    \includegraphics[width=\linewidth]{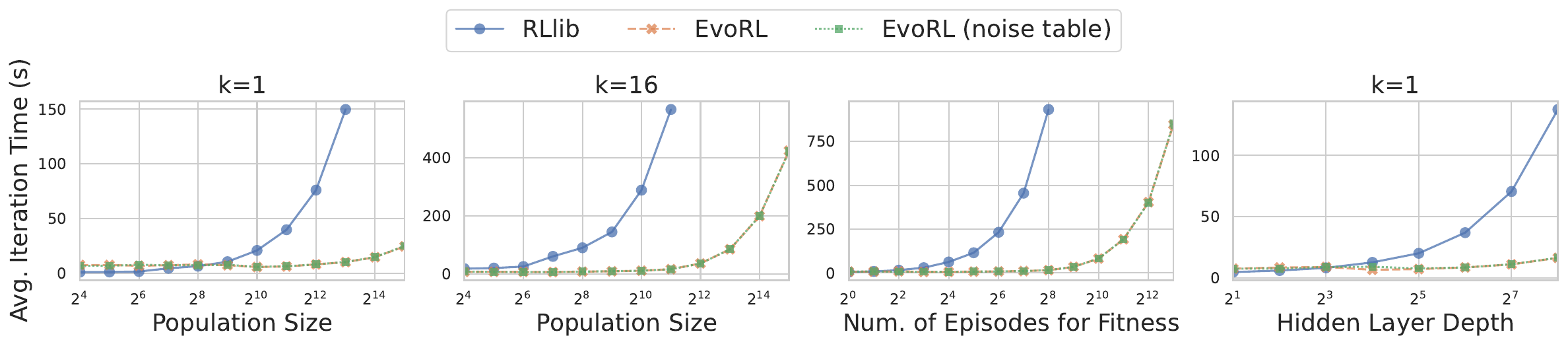}
    \caption{The average iteration time of OpenES across different implementations. RLlib utilizes a distributed pipeline on CPUs, whereas ours employs an end-to-end pipeline on a single GPU, achieving a maximum speed-up of over 60x in the RLlib range.}
    \label{fig:es-perf}
\end{figure}

\begin{figure}[t]
    \centering
    \begin{subfigure}[t]{0.3\linewidth}
        \centering
        \includegraphics[width=\linewidth]{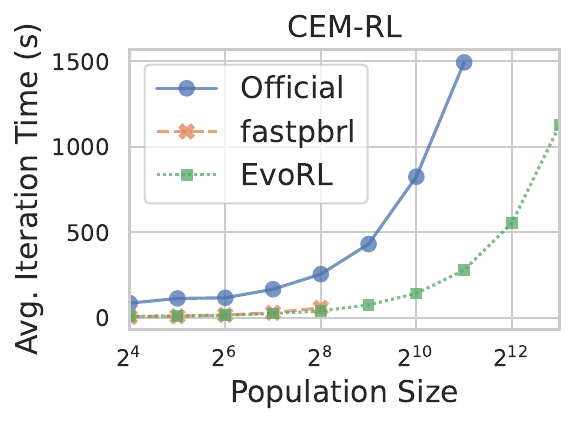}
        \caption{CEM-RL}
        \label{fig:erl-perf}
    \end{subfigure}
    \hfill
    \begin{subfigure}[t]{0.6\linewidth}
        \centering
        \includegraphics[width=\linewidth]{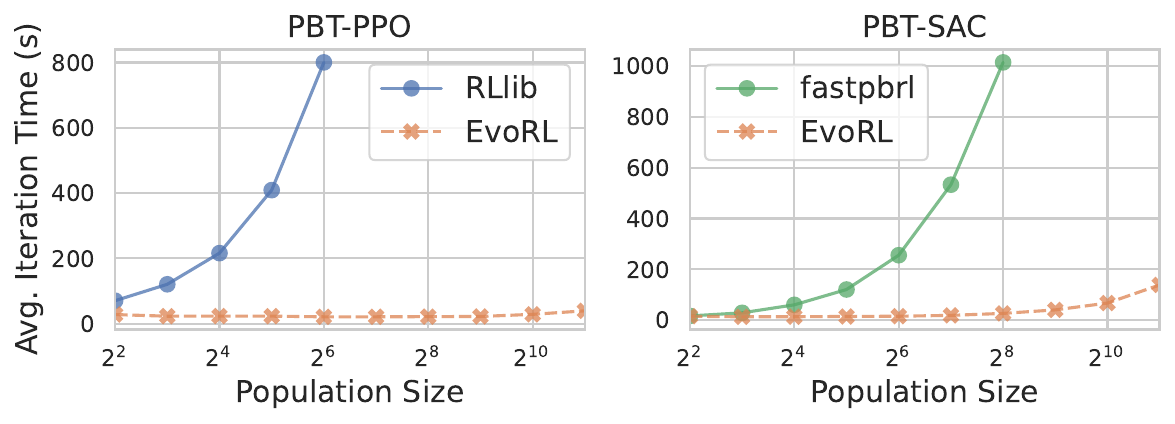}
        \caption{PBT}
        \label{fig:pbt-perf}
    \end{subfigure}
    \caption{The average iteration time of \textbf{(a) CEM-RL} and \textbf{(b) PBT} in different implementations. For CEM-RL, the fastpbrl implementation encounters out-of-memory issues when the population size exceeds 256, whereas our implementation achieves a 5–9x speed-up over the official implementation and a peak speed-up of 1.45x over fastpbrl. For PBT, we employ PPO as the underlying on-policy RL algorithm and SAC as the underlying off-policy RL algorithm. Due to memory constraints, the population sizes for PBT in RLlib and fastpbrl are also limited. Under these conditions, our implementation achieves a maximum speed-up of 30–40x.}
\end{figure}

We evaluate the performance of \evorl{}’s OpenES implementation against the RLlib implementation\footnote{\url{https://docs.ray.io/en/latest/rllib}}, which uses the CPU-based distributed framework Ray. Their implementation uses a full CPU-based architecture, including rollout and agent update. Besides, in RLlib, a large shared noise table is pre-built at the start of training to efficiently generate noise for new populations \citep{salimansEvolutionStrategiesScalable2017}. In contrast, \evorl{}’s default implementation generates noise on-the-fly during training. To ensure fair comparisons, we also test a variant of \evorl{} that incorporates a pre-built noise table.

We train a deterministic policy with $m$ hidden layers of size 256 and set the population size to $n$. The number of episodes for fitness evaluation is $k$. The base configuration is $m=2$, $n=128$, $k=1$, and experiments are conducted by varying one parameter at a time to analyze scalability across different dimensions (e.g., increasing $m$, $n$, or $k$). Results, presented in Fig.~\ref{fig:es-perf}, show that \evorl{}’s on-the-fly noise generation imposes negligible performance overhead due to the efficiency of GPU-based computations. Moreover, \evorl{} achieves a maximum speed-up of over 60x compared to RLlib under various settings, underscoring the efficiency of its end-to-end GPU-accelerated pipeline.

\subsubsection{Performance of ERL Algorithms}
\label{sec:speed-exp-erl}

We compare the performance of CEM-RL as implemented in \evorl{} with two existing implementations: the original Python-based implementation\footnote{\url{https://github.com/apourchot/CEM-RL}} and the fastpbrl implementation\footnote{\url{https://github.com/instadeepai/fastpbrl}}. The original implementation sequentially performs rollouts and gradient updates for each agent using Python for-loops, leading to inefficient training. The fastpbrl implementation improves upon this by leveraging JAX vectorization for RL updates on GPUs. However, its environments are CPU-based, and rollouts are parallelized across multiple CPU cores using Python’s \lstinline|multiprocessing|. This architecture introduces additional CPU-GPU communication overhead from agent weights and trajectories transfer, especially for large population sizes.

As shown in Fig.~\ref{fig:erl-perf}, \evorl{}’s fully GPU-accelerated pipeline eliminates this communication overhead by performing rollouts, RL updates, and environment simulations entirely on the GPU. This design achieves a 5–9x speed-up compared to the official implementation and a 1.45x speed-up to fastpbrl before its out-of-memory issues. \evorl{} demonstrates robust scalability with increased population sizes, further emphasizing its efficiency in large-scale ERL tasks.

\subsubsection{Performance of Population-based AutoRL Algorithms}
\label{sec:speed-exp-pbt}

We compare \evorl{}’s PBT implementation with both the Ray implementation\footnote{\url{https://docs.ray.io/en/latest/tune/}}, which tunes PPO hyperparameters, and the fastpbrl PBT implementation, which tunes SAC hyperparameters. The Ray implementation employs a hybrid architecture where multiple PPO training instances are parallelized across CPU processes and share a single GPU for updates. Similarly, fastpbrl parallelizes rollouts via multiprocessing and uses JAX-based vectorization for parallel RL updates. Besides, its shared replay buffer is stored on CPU memory, resulting in frequent data transfers between CPU and GPU.

In contrast, \evorl{} executes the entire PBT pipeline on the GPU, including rollouts, RL updates, and a GPU VRAM-based shared replay buffer. Results, shown in Fig.~\ref{fig:pbt-perf}, highlight that \evorl{} achieves a maximum speed-up of 30-40x compared to both Ray and fastpbrl implementations. Furthermore, \evorl{} avoids memory limitations observed in competing frameworks, supporting larger populations and more computationally intensive configurations.

In summary, by leveraging hierarchical vectorization on the fully GPU-accelerated pipelines, \evorl{} achieves significant speed-ups and scalability improvements across all tested algorithms, including OpenES, CEM-RL, and PBT. These results demonstrate \evorl{}’s capacity to efficiently handle large-scale training configurations.

\begin{figure}[t]
    \centering
    \includegraphics[width=\linewidth]{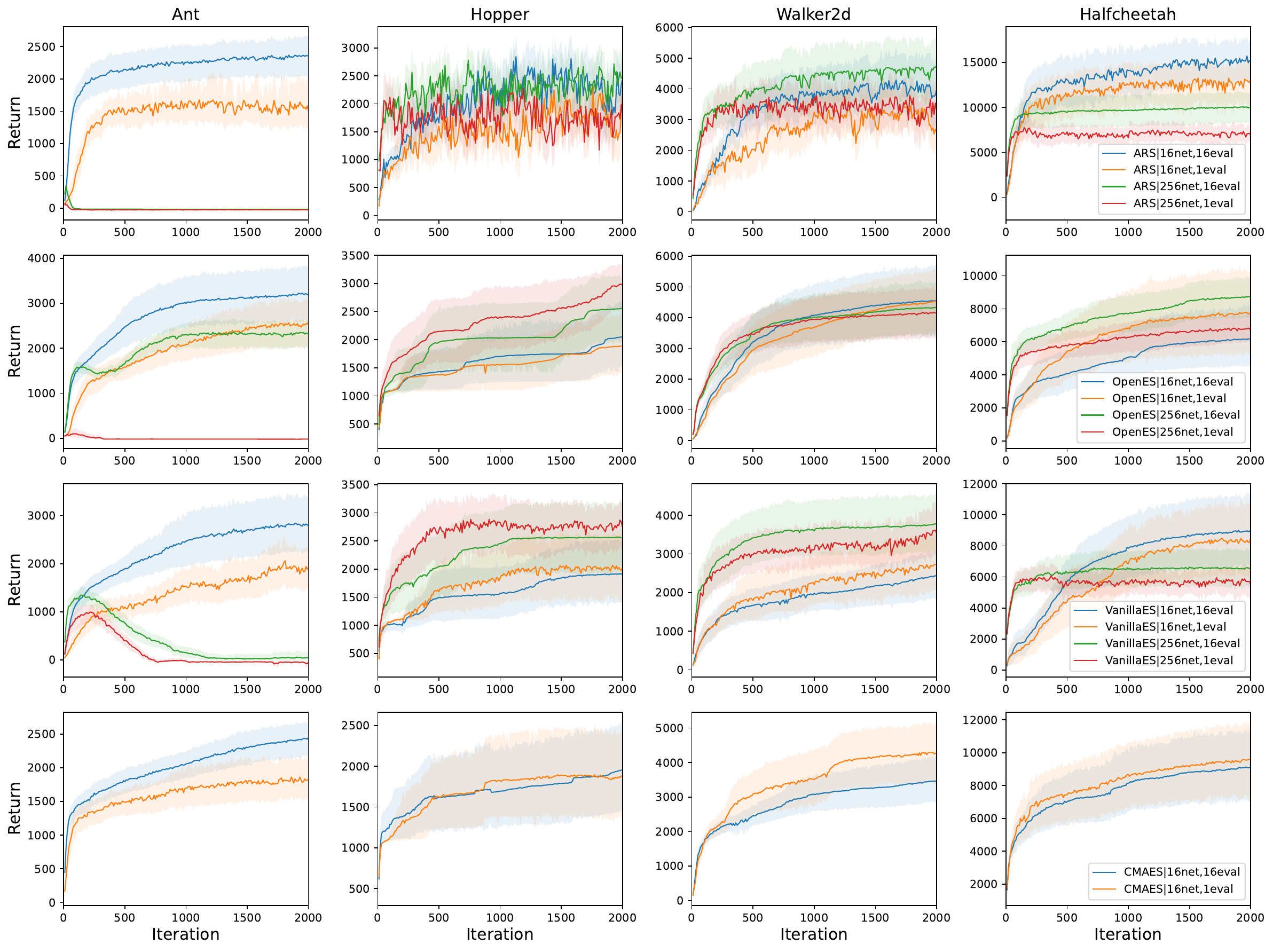}
    \caption{The benchmark of different evolutionary algorithms on robotic locomotion tasks. We compare the performance of algorithms with $[16,16]$ or $[256,256]$ hidden layers (\textbf{16net} or \textbf{256net}) and 1 or 16 episodes for fitness evaluation (\textbf{1eval} or \textbf{16eval}). Each configuration is repeated with 16 different seeds. The average return and its 95\% confidence interval about the population distribution mean are reported.}
    \label{fig:es-results}
\end{figure}

\subsection{Use Cases}
\label{sec:use-cases}

The versatility and modular design of \evorl{} enable it to support a wide range of algorithms, making it a valuable platform for EvoRL research. 
In this section, we demonstrate the applicability of \evorl{} through three key use cases, showcasing \evorl{}’s ability to handle diverse algorithmic paradigms on the GPU-accelerated pipelines.

\subsubsection{Use Case of EAs for Policy Search}
\label{sec:ea-exp}

We evaluate the performance of modern EAs for policy search, including CMA-ES \citep{hansenCMAEvolutionStrategy2016}, OpenES \citep{salimansEvolutionStrategiesScalable2017}, VanillaES \citep{chrabaszczBackBasicsBenchmarking2018}, and ARS \citep{maniaSimpleRandomSearch2018}. 
The experiments are conducted with a fixed population size of 128, where each individual represents an agent parameterized by a deterministic policy network with a two-layer MLP architecture. 
The fitness of an agent is measured as the average undiscounted return over multiple episodes without action exploration. 
To measure the population quality, we independently compute the average return of the agent during training from 128 episodes, whose weights correspond to the mean of the population distribution.

To investigate the impact of network size and noise in fitness evaluation, we test two hidden layer configurations, $[16,16]$ and $[256,256]$, and vary the number of episodes used for fitness evaluation (1 or 16). All experiments run for 2000 iterations, and each configuration is repeated 16 times with different random seeds to ensure statistical robustness.
Due to the high memory requirements of CMA-ES, experiments involving $[256,256]$ networks are omitted, as maintaining the covariance matrix exceeds the available GPU memory.

As presented in Fig.~\ref{fig:es-results}, the results reveal that larger networks tend to achieve faster convergence but do not necessarily lead to superior final performance. 
For example, performance degradation is observed with $[256,256]$ networks in the Ant environment. 
Similarly, in the HalfCheetah environment, $[256,256]$ networks yield lower returns than $[16,16]$ networks when using ARS and VanillaES. 
Increasing the number of episodes for fitness evaluation improves the accuracy of fitness estimates in noisy environments. 
However, this improvement might come at the cost of reduced exploration by fitness noise, which may result in suboptimal solutions in certain cases. These findings highlight the trade-offs between network size and evaluation noise in evolutionary policy search.
In addition, Table.~\ref{tab:ea-benchmark} provides the best average return across different EAs on these environments. ARS and OpenES are two advanced EAs, achieving better performance for policy search.

\begin{table}
    \centering
    \caption{Performance of different EAs on robotic locomotion tasks. The best average return across 16 runs are reported under 4 configurations --- \textcolor{16n16e}{Blue}: 16net,16eval; \textcolor{16n1e}{Orange}: 16net,1eval; \textcolor{256n16e}{Green}: 256net,16eval; \textcolor{256n1e}{Red}: 256net,1eval. The highest return for each environment is highlighted.}
    \label{tab:ea-benchmark}
    \begin{tabular}{*{5}{c}}
        \toprule
        Environment                  & ARS                                   & OpenES                               & VanillaES                    & CMA-ES                      \\ \midrule
        \multirow{4}{*}{Ant}         & \textcolor{16n16e}{2368.85}           & \textcolor{16n16e}{\underline{\textbf{3224.73}}} & \textcolor{16n16e}{2838.81}  & \textcolor{16n16e}{2449.74} \\
                                     & \textcolor{16n1e}{1675.04}            & \textcolor{16n1e}{2581.69}           & \textcolor{16n1e}{2060.51}   & \textcolor{16n1e}{1863.62}  \\
                                     & \textcolor{256n16e}{349.61}           & \textcolor{256n16e}{2357.23}         & \textcolor{256n16e}{1347.56} & -                           \\
                                     & \textcolor{256n1e}{68.19}             & \textcolor{256n1e}{100.50}           & \textcolor{256n1e}{995.85}   & -                           \\ \midrule
        \multirow{4}{*}{Hopper}      & \textcolor{16n16e}{2848.12}           & \textcolor{16n16e}{2046.94}          & \textcolor{16n16e}{1920.72}  & \textcolor{16n16e}{1958.89} \\
                                     & \textcolor{16n1e}{2333.21}            & \textcolor{16n1e}{1894.04}           & \textcolor{16n1e}{2071.13}   & \textcolor{16n1e}{1896.59}  \\
                                     & \textcolor{256n16e}{2786.01}          & \textcolor{256n16e}{2561.93}         & \textcolor{256n16e}{2569.93} & -                           \\
                                     & \textcolor{256n1e}{2282.49}           & \textcolor{256n1e}{\underline{\textbf{2987.82}}} & \textcolor{256n1e}{2884.96}  & -                           \\ \midrule
        \multirow{4}{*}{Walker2d}    & \textcolor{16n16e}{4288.57}           & \textcolor{16n16e}{4567.66}          & \textcolor{16n16e}{2442.28}  & \textcolor{16n16e}{3465.87} \\
                                     & \textcolor{16n1e}{3493.26}            & \textcolor{16n1e}{4544.91}           & \textcolor{16n1e}{2743.47}   & \textcolor{16n1e}{4300.73}  \\
                                     & \textcolor{256n16e}{\underline{\textbf{4755.73}}} & \textcolor{256n16e}{4332.87}         & \textcolor{256n16e}{3780.70} & -                           \\
                                     & \textcolor{256n1e}{3786.71}           & \textcolor{256n1e}{4172.71}          & \textcolor{256n1e}{3615.59}  & -                           \\ \midrule
        \multirow{4}{*}{HalfCheetah} & \textcolor{16n16e}{\underline{\textbf{15727.93}}} & \textcolor{16n16e}{6183.75}          & \textcolor{16n16e}{9028.43}  & \textcolor{16n16e}{9122.02} \\
                                     & \textcolor{16n1e}{13265.43}           & \textcolor{16n1e}{7793.73}           & \textcolor{16n1e}{8514.64}   & \textcolor{16n1e}{9594.04}  \\
                                     & \textcolor{256n16e}{10059.23}         & \textcolor{256n16e}{8747.80}         & \textcolor{256n16e}{6637.56} & -                           \\
                                     & \textcolor{256n1e}{7750.28}           & \textcolor{256n1e}{6831.58}          & \textcolor{256n1e}{6073.02}  & -                           \\ \bottomrule
    \end{tabular}
\end{table}

\subsubsection{Use Case of ERL Algorithms}
\label{sec:erl-exp}

We implement the original ERL and CEM-RL algorithms, incorporating modifications proposed by \citet{flajoletFastPopulationBasedReinforcement2022}. 
These modifications replace sequential population updates with parallel vectorized updates, significantly improving computational efficiency without compromising algorithm performance. Both ERL and CEM-RL rely on off-policy RL methods and replay buffers to store and sample experiences. 
While the importance of replay ratios, i.e., the number of gradient updates per environment step, has been extensively studied in standard off-policy RL algorithms \citep{fedusRevisitingFundamentalsExperience2020}, it remains an underexplored area in the context of ERL.

The original implementations dynamically scale the number of RL updates per iteration based on the total number of timesteps collected -- either from the current iteration for CEM-RL or from all iterations for ERL. However, dynamically scaling updates with cumulative timesteps results in escalating computational costs in later iterations. To address this, we modify ERL to align the number of RL updates per iteration with the timesteps sampled from the current iteration, similar to CEM-RL. Additionally, we introduce a fixed number of RL updates per iteration, decoupling the updates from population size and agent performance (e.g., episode length). This adjustment significantly reduces computational overhead, particularly for larger population sizes, while maintaining scalability.

For our experiments, we use a population size of 10, consistent with the original implementations, and terminate training after 20,000 episodes. All CEM-RL experiments were initialized with 25,600 random timesteps in the replay buffer to ensure stable learning in the early training phases.
The fixed number of RL updates per iteration is set to 4,096 for most environments. However, significant performance degradation was observed in the Ant environment with this configuration. To mitigate this, we reduced the number of updates to 1,024 per iteration for both ERL and CEM-RL and added 25,600 random timesteps for ERL.

\begin{figure}
    \centering
    \includegraphics[width=\linewidth]{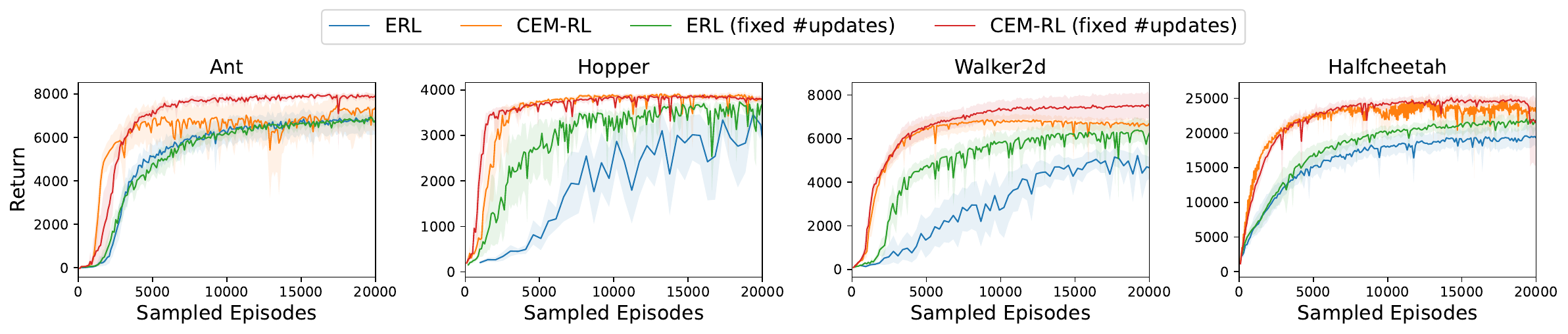}
    \caption{The benchmark of ERL and CEM-RL on robotic locomotion tasks. Each algorithm is repeated with 8 different seeds. The average return and its 95\% confidence interval about the population distribution mean are reported.}
    \label{fig:erl-results}
\end{figure}

Fig.~\ref{fig:erl-results} illustrates the comparative performance of ERL and CEM-RL, as well as the impact of using a fixed number of RL updates. CEM-RL consistently achieves faster convergence and outperforms ERL across all tested environments. Table~\ref{tab:erl-rl-updates} summarizes the total number of RL updates during training, revealing that using a fixed number of updates typically reduces the total computational cost. This reduction is particularly pronounced in CEM-RL, where the population maintains high-quality agents, leading to more timesteps sampled per iteration. 

These results suggest that using a fixed number of RL updates per iteration achieves comparable or even superior performance to the original implementation while substantially reducing training costs. We hypothesize that fixed RL updates allocate more updates in the early training stage, thus allowing agents to acquire foundational skills quickly. This finding also explains the need for additional random timesteps at the start of training in this configuration, which helps improve the exploration diversity.

\begin{table}
    \centering
    \caption{The total number of RL updates during the training of ERL and CEM-RL.}
    \begin{tabular}{*{5}{c}}
        \toprule
        Algorithms  & Ant    & Hopper & Walker2d & HalfCheetah \\ \midrule
        ERL         & 8.27M  & 7.59M  & 5.46M    & 20.00M      \\
        ERL (fixed \#updates)    & 1.86M  & 7.45M  & 7.45M    & 7.45M       \\
        CEM-RL      & 15.15M & 16.38M & 17.78M   & 20.00M      \\
        CEM-RL (fixed \#updates) & 2.05M  & 8.19M  & 8.19M    & 8.19M       \\ \bottomrule
    \end{tabular}
    \label{tab:erl-rl-updates}
\end{table}

\subsubsection{Use Case of Population-based AutoRL}
\label{sec:pbt-exp}

Population-Based Training (PBT) is a representative population-based AutoRL algorithm that dynamically adjusts the hyperparameters of underlying learning algorithms during training. 
PBT employs an exploit-and-explore strategy, where the top-performing 20\% of instances are used to replace the bottom 20\% based on a predefined meta-objective, forming the exploitation phase. 
This replacement involves both the internal states of the algorithm, such as agent weights, and the associated hyperparameters. 
Subsequently, the replaced bottom instances undergo hyperparameter perturbation during the exploration phase. 
From an EC perspective, PBT’s exploit-and-explore strategy corresponds to selection and mutation operators, respectively.

In this work, we extend the standard PBT framework by introducing an abstract evolutionary layer that generalizes its exploit-and-explore strategy. 
We evaluate this extension using a modified variant, {PBT-CSO}, which incorporates Competitive Swarm Optimization (CSO) operators \citep{chengCompetitiveSwarmOptimizer2015} into the evolutionary layer. 
PBT-CSO can be regarded as a synchronized variant of GPBT-PL \citep{baiGeneralizedPopulationBasedTraining2024}, leveraging CSO-inspired dynamics for hyperparameter updates. 
Specifically, PBT-CSO pairs individuals in the population randomly, forming $\{(\theta_i, \theta_j)\}_{n/2}$ pairs, where $n$ denotes the population size. 
Each individual $\theta_i$ includes both the internal algorithm state $\theta_i^w$ and the associated hyperparameters $\theta_i^h$. 
For each pair, a {student} $\theta_s$ and a {teacher} $\theta_t$ are identified based on their performance with respect to the meta-objective. 
The student’s internal state $\theta_s^w$ is replaced by the teacher’s state $\theta_t^w$, and the student’s hyperparameters $\theta_s^h$ are updated using CSO-inspired equations:
\[
v_s := r_1 v_s + r_2 (\theta_t^h - \theta_s^h), \quad \theta_s^h := \theta_s^h + v_s,
\]
where $v_s$ represents the velocity of the student, initialized to zero for all individuals at the start of training, and $r_1, r_2$ are random values sampled uniformly from $[0,1]$.

\begin{figure}
    \centering
    \includegraphics[width=\linewidth]{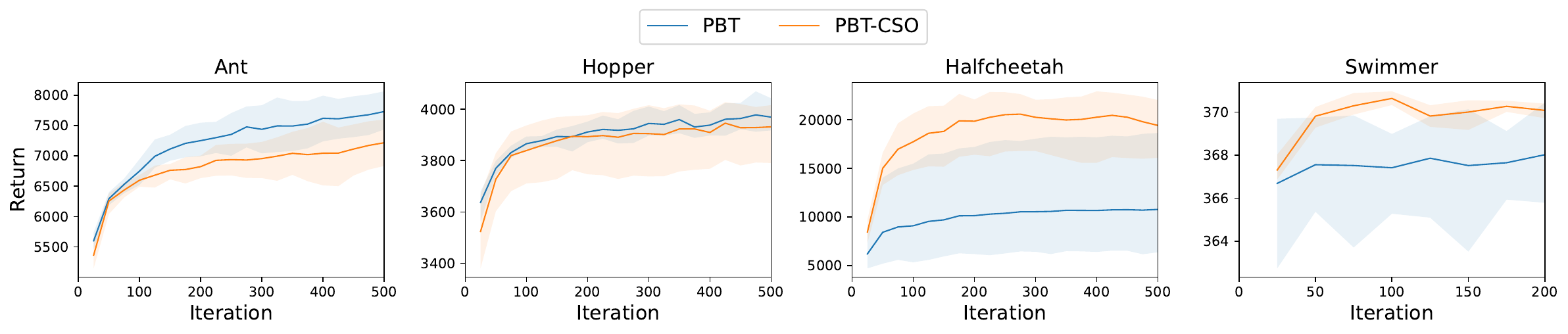}
    \caption{The benchmark of PBT and PBT-CSO on PPO hyperparameter tuning. Each algorithm is repeated with 4 different seeds. The average best return of the population and 95\% confidence interval are reported.}
    \label{fig:pbt-results}
\end{figure}

We conduct experiments comparing the original PBT with PBT-CSO, using 128 training instances for hyperparameter tuning on PPO. Fig.~\ref{fig:pbt-results} presents the performance comparison, highlighting that PBT-CSO consistently outperforms the original PBT in specific environments. 
The incorporation of CSO-inspired operators allows for more refined hyperparameter updates, resulting in improved training performance.
Overall, our results demonstrate the benefits of extending PBT with an abstract evolutionary layer, as seen in the superior performance of PBT-CSO. This approach opens new possibilities for enhancing PBT by exploring diverse algorithms in EC, making it a promising direction for future research in AutoRL.



\section{Related Work}

Despite the promising results of EvoRL algorithms, the field lacks efficient and scalable frameworks to fully exploit their potential. 
Existing implementations are often developed independently, without a unified interface, and may suffer from inefficiencies. 
In this section, we summarize prior libraries that cover some functionalities provided by \evorl{} and highlight their limitations.

\subsection*{Reinforcement Learning Libraries in JAX}

Several JAX-based RL libraries have been developed to leverage the advantages of JAX’s high-performance computing capabilities. Examples include rlax \citep{deepmind2020jax}, dopamine \citep{castroDopamineResearchFramework2018}, PureJaxRL \citep{luDiscoveredPolicyOptimisation2022}, Mava \citep{dekock2023mava}, and Stoix \citep{dekock2023mava}. These libraries achieve significant speedups by integrating JAX-based environments for end-to-end training on accelerators. However, they are primarily focused on traditional RL methods and lack support for evolutionary computation (EC) components, which are crucial for EvoRL.

\subsection*{Evolutionary Computation Libraries in JAX}

Recent JAX-based libraries for EC have also shown promising results, such as EvoJax \citep{tangEvoJAXHardwareAcceleratedNeuroevolution2022}, evosax \citep{langeEvosaxJAXbasedEvolution2022}, and EvoX \citep{huangEvoXDistributedGPUaccelerated2023}. These libraries provide robust tools for general optimization tasks and demonstrate excellent performance on JAX-enabled accelerators. 
However, they lack RL-specific components, such as unified environment interfaces, observation normalization, and other utilities required for policy search tasks. 
In contrast, \evorl{} bridges this gap by offering rich modular components that integrate EC pipelines with RL environments, enabling seamless construction of policy search algorithms.

\subsection*{EvoRL-Specific Libraries}

Limited libraries have explored EvoRL-specific functionalities. Examples include RLlib \citep{liangRLlibAbstractionsDistributed2018} and fastpbrl \citep{flajoletFastPopulationBasedReinforcement2022}, which support PBT for hyperparameter optimization. fastpbrl also implements a few ERL algorithms, such as CEM-RL \citep{pourchotCEMRLCombiningEvolutionary2019} and DvD \citep{parker-holderEffectiveDiversityPopulation2020}. Similarly, Lamarckian \citep{baiLamarckianPlatformPushing2022} provides a platform for EvoRL, including EAs and PBT algorithms; and Pearl \citep{tangri2022pearl} provides EC and ERL supports.
However, these libraries are designed for CPU-based environments and rely on hybrid pipeline architectures to orchestrate heterogeneous devices. For example, RLlib and fastpbrl use multiprocessing to distribute rollout processes across multiple CPU cores, requiring manual asynchronous scheduling and signal control to manage the training pipeline efficiently. These architectures increase complexity and introduce communication overhead between CPUs and GPUs.

In contrast, \evorl{} eliminates these limitations by offering a direct, end-to-end training pipeline entirely on homogeneous devices like GPUs. This design removes the need for multiprocessing and CPU-GPU communication tricks, significantly improving the efficiency of EvoRL algorithms, even on a single GPU. The streamlined architecture allows researchers to focus on algorithm design and implement training logic more intuitively.

\subsection*{Closest Related Libraries}

The most closely related libraries to \evorl{} are QDax \citep{chalumeauQDaxLibraryQualityDiversity2024} and PBRL \citep{shahidScalingPopulationBasedReinforcement2024}, which also support end-to-end training for EvoRL algorithms. However, QDax is primarily tailored to quality diversity (QD) algorithms, while \evorl{} is designed to support a broader range of EvoRL algorithms, including ERL and Population-Based AutoRL. PBRL focuses on PBT pipelines for Isaac Gym environments, whereas \evorl{} offers compatibility with a wider variety of environments and provides a richer set of EC modules. These features make \evorl{} a more versatile and comprehensive platform for EvoRL research.

\section{Conclusion}

Evolutionary Reinforcement Learning (EvoRL) has emerged as a promising paradigm, offering the benefits of population-based exploration and gradient-free optimization to address the challenges faced by traditional reinforcement learning (RL). 
Simultaneously, advancements in GPU-accelerated environments and computing frameworks have created new opportunities for scalable and efficient algorithm development. 
Recognizing the need for a unified, high-performance platform to support EvoRL research, we introduced \evorl{}, a comprehensive framework designed to fully exploit the capabilities of modern accelerators.

\evorl{} achieves end-to-end GPU acceleration for the entire EvoRL pipeline, encompassing environment simulations, evolutionary algorithms, and RL components. 
By leveraging hierarchical vectorization and JAX-based optimization techniques, \evorl{} delivers significant improvements in computational efficiency and scalability, as demonstrated in our experiments. 
The framework eliminates the bottlenecks of CPU-GPU communication and reduces the complexities associated with traditional multiprocessing-based implementations, enabling seamless training of large-scale populations on a single GPU. These advancements lower the computational barriers to EvoRL research and facilitate large-scale experimentation.

In addition to its performance-oriented design, \evorl{} offers a modular and extensible architecture that simplifies the integration and customization of various algorithmic components.
Researchers can easily explore new hybrid EvoRL paradigms, benchmark algorithms, and conduct ablation studies within a standardized and user-friendly interface. The support for diverse use cases, including Evolution-guided RL (ERL), Population-Based Training (PBT), and canonical evolutionary algorithms (EAs), further highlights \evorl{}’s versatility.

By providing a scalable, efficient, and accessible platform, \evorl{} addresses critical limitations in existing frameworks and paves the way for the next generation of EvoRL research. 
As GPU-accelerated environments continue to evolve, \evorl{} is well-positioned to drive innovation in the field, fostering the development of novel algorithms and facilitating their application to increasingly complex and demanding tasks.

\begin{acks}
    This work was supported in part by Guangdong Basic and Applied Basic Research Foundation (No. 2024B1515020019). 
\end{acks}

\bibliographystyle{ACM-Reference-Format}
\bibliography{evorl}

\newpage
\appendix

\section{Example Codes}
\label{sec:example-codes}

\evorl{} follows the object-oriented functional programming style, where classes define the running logic, and their state is externally maintained. It helps reuse components in single or batch mode. We provide an example of how the Agent and Workflow of A2C are defined and used in \evorl{}.

\begin{lstlisting}[]
class A2CAgent(Agent):
    continuous_action: bool
    policy_network: nn.Module
    value_network: nn.Module
    obs_preprocessor: Any = pytree_field(default=None, static=True)

    def init(self, obs_space, action_space, key):
        ...
    
    def compute_actions(self, agent_state, sample_batch, key):
        ...
    
    def evaluate_actions(self, agent_state, sample_batch, key):
        ...
    
    def loss(self, agent_state, sample_batch, key):
        ...

policy_network = make_policy_network()
value_network = make_v_network()

key = jax.random.PRNGKey(42)
agent = A2CAgent(False, policy_network, value_network)
agent_state = agent.init(env.obs_space, env.action_space, key)
\end{lstlisting}
\begin{lstlisting}[]
class A2CWorkflow(OnPolicyWorkflow):
    @classmethod
    def _build_from_config(cls, config: DictConfig):
        env = create_env(config.env, ...)
        agent = A2CAgent(...)
        optimizer = optax.adam(config.optimizer.lr)
        eval_env = create_env(config.env, ...)
        evaluator = Evaluator(...)

        return cls(env, agent, optimizer, evaluator, config)

    def step(self, state):
        trajectory, env_state = rollout(
            state.env_state, state.agent_state, rollout_length=self.config.rollout_length
        )
        loss, agent_state, opt_state = update_fn(
            state.opt_state, state.agent_state, trajectory, learn_key
        )
        ...

        state = state.replace(
            ...
            agent_state=agent_state,
            env_state=env_state,
            opt_state=opt_state,
        )

        return train_metrics, state

    def evaluate(self, state):
        eval_metrics = self.evaluator.evaluate(
            state.agent_state, eval_key, num_episodes=self.config.eval_episodes
        )
        ...

        return eval_metrics, state

    def learn(self, state):
        for i in range(num_iters):
            train_metrics, state = self.step(state)
            self.recorder.write(train_metrics, i)

            if i % self.config.eval_interval == 0:
                eval_metrics, state = self.evaluate(state)
                self.recorder.write(eval_metrics, i)

            self.checkpoint_manager.save(i, ...)

        return state

key = jax.random.PRNGKey(42)
workflow = A2CWorkflow.build_from_config(config)
state = workflow.init(key)
state = workflow.learn(state)
\end{lstlisting}

In the example, the \lstinline|A2CAgent| defines how the actions are computed when applying rollout at the training and evaluation stage by \lstinline|compute_actions()| and \lstinline|evaluate_actions()|, respectively. The agent is agnostic to the network structures of the actor and critic, which are passed as arguments at the creation stage of the agent. Then \lstinline|agent.init(key)| is called to get the state of the agent \lstinline|agent_state|, which includes the networks' weights and other persistent data such as \lstinline|obs_preprocessor_state| for observation normalization. The \lstinline|agent_state| is then used for computing actions and losses.

The \lstinline|A2CWorkflow| defines the training pipeline of A2C. Its \lstinline|_build_from_config()| defines the fundamental components in \lstinline|OnPolicyWorkflow|, including the rollout environment, the agent object (\lstinline|A2CAgent|) and its optimizer, the independent evaluator, and the config object during the creation stage. This method will be called when the workflow is created by its public method: \lstinline|A2CWorkflow.build_from_config(config)|.
The components' states, along with the workflow's own states, will be initialized in \lstinline|init()|.
Its \lstinline|step()| defines one iteration of the training. For A2C, it includes: (1) sampling trajectories from environments, (2) using these trajectories as data batches to compute the loss and apply gradient-based updates for \lstinline|agent_state|. The \lstinline|step()| function can be JIT compiled and vectorized by JAX. Moreover, \lstinline|learn()| defines the training loop, which calls \lstinline|step()| multiple times and handles the termination condition, evaluation, and stuff that cannot be JIT compiled, like logging and checkpointing. The usage of \lstinline|A2CWorkflow| also follows the object-oriented functional programming style, where its state is separately initialized and externally stored after the object instantiation.

\section{Additional Details about Training Settings}
\label{sec:training-settings}

\subsection{Evolutionary Algorithms for Policy Search}

We apply the observation normalization when training different EAs. We choose virtual batch normalization \citep{salimansEvolutionStrategiesScalable2017}, which computes the mean and standard deviation of the observations from 10000 random timesteps and keeps them fixed for the entire training process. For ARS, we additionally follow the original implementation \citep{maniaSimpleRandomSearch2018}, which dynamically updates these statistics from newly collected observations (denoted as RS). We also use ReLU as the activation function of MLPs.
The algorithm-specific hyperparameters follow their original implementations and are fixed across different environments, as listed in Table~\ref{tab:ea-hp}. 2000 iterations are executed for each EA.
For experiments in Sec.~\ref{sec:speed-exp-ea}, we record the average iteration time from 5 iterations of different implementations.

\begin{table}
    \centering
    \caption{Hyperparameters for Evolutionary Algorithms}
    \label{tab:ea-hp}
    \begin{subtable}[b]{0.24\textwidth}
        \centering
        \resizebox{\linewidth}{!}{
            \begin{tabular}{cc}
                \toprule
                Hyperparameter  & Value \\ \midrule
                Population Size & 128   \\
                Num. Elites     & 16    \\
                Optimizer       & SGD   \\
                Learning Rate   & 0.02  \\
                Noise Std.      & 0.03  \\
                Obs. Norm.      & RS    \\ \bottomrule
            \end{tabular}
        }
        \caption{ARS}
    \end{subtable}
    \begin{subtable}[b]{0.24\textwidth}
        \centering
        \resizebox{\linewidth}{!}{
            \begin{tabular}{cc}
                \toprule
                Hyperparameter  & Value \\ \midrule
                Population Size & 128   \\
                Optimizer       & Adam  \\
                Learning Rate   & 0.01  \\
                Weight Decay    & 0.005 \\
                Noise Std.      & 0.02  \\
                Obs. Norm.      & VBN   \\ \bottomrule
            \end{tabular}
        }
        \caption{OpenES}
    \end{subtable}
    \begin{subtable}[b]{0.24\textwidth}
        \centering
        \resizebox{\linewidth}{!}{
            \begin{tabular}{cc}
                \toprule
                Hyperparameter  & Value \\ \midrule
                Population Size & 128   \\
                Num. Elites     & 16    \\
                Noise Std.      & 0.02  \\
                Obs. Norm.      & VBN   \\ \bottomrule
            \end{tabular}
        }
        \caption{VanillaES}
    \end{subtable}
    \begin{subtable}[b]{0.24\textwidth}
        \centering
        \resizebox{\linewidth}{!}{
            \begin{tabular}{cc}
                \toprule
                Hyperparameter  & Value \\ \midrule
                Population Size & 128   \\
                Num. Elites     & 64    \\
                Init Noise Std. & 0.1   \\
                Obs. Norm.      & VBN   \\ \bottomrule
            \end{tabular}
        }
        \caption{CMA-ES}
    \end{subtable}
\end{table}

\subsection{ERL Algorithms}

For the original ERL and CEM-RL, we use $[256,256]$ MLP networks for actors and a shared critic. ReLU and layer normalization are applied to the hidden layers, aligning with the original implementations.
The underlying RL algorithm is TD3, and no observation normalization is applied. All ERL algorithms are trained with 20000 sampled episodes.
Since the RL update is applied on a shared critic, we don't delay the actors' update (i.e., the actor update interval is one), which is the same as the implementation in fastpbrl. Table~\ref{tab:erl-hp} lists the hyperparameters for ERL algorithms and the underlying TD3 algorithm.
To test the speed in Sec.~\ref{sec:speed-exp-erl}, we choose fixed 4096 RL updates per iteration and record the average iteration time from 5 iterations for different implementations.
The other training settings are the same as above, and the time spent on the warm-up stage is skipped.

\begin{table}
    \centering
    \caption{Hyperparameters for ERL Algorithms}
    \label{tab:erl-hp}
    \begin{subtable}[b]{0.3\textwidth}
        \centering
        \resizebox{\linewidth}{!}{
            \begin{tabular}{cc}
                \toprule
                Hyperparameter       & Value \\ \midrule
                Population Size      & 10    \\
                Num. RL Agents       & 1     \\
                Random Timesteps     & 0     \\
                Episodes for Fitness & 1     \\
                Warm-up Iters        & 10    \\
                Evaluation Episodes  & 128   \\
                Replay Buffer Size   & 1e6   \\ \bottomrule
            \end{tabular}
        }
        \caption{ERL}
    \end{subtable}
    \begin{subtable}[b]{0.33\textwidth}
        \centering
        \resizebox{\linewidth}{!}{
            \begin{tabular}{cc}
                \toprule
                Hyperparameter       & Value        \\ \midrule
                Population Size      & 10           \\
                Num. Elites          & 5            \\
                Num. RL Agents       & 5            \\
                Noise Std.           & 1e-3 to 1e-5 \\
                Random Timesteps     & 25600        \\
                Episodes for Fitness & 1            \\
                Warm-up Iters        & 10           \\
                Evaluation Episodes  & 128          \\
                Replay Buffer Size   & 1e6          \\ \bottomrule
            \end{tabular}
        }
        \caption{CEM-RL}
    \end{subtable}
    \begin{subtable}[b]{0.33\textwidth}
        \centering
        \resizebox{\linewidth}{!}{
            \begin{tabular}{cc}
                \toprule
                Hyperparameter        & Value \\ \midrule
                Discount Factor       & 0.99  \\
                Soft Update ratio     & 0.005 \\
                Exploration Epsilon   & 0.1   \\
                Policy Noise          & 0.2   \\
                Action Noise Clip     & 0.5   \\
                Batch Size            & 256   \\
                Optimizer             & Adam  \\
                Learning Rate         & 3e-4  \\
                Actor Update Interval & 1     \\ \bottomrule
            \end{tabular}
        }
        \caption{Underlying TD3}
    \end{subtable}
\end{table}

\subsection{Population-based Training}

For PBT and PBT-CSO, the underlying RL algorithm is PPO. We use $[256,256]$ MLP networks for its actor and critic. The hyperparameters for PBT and PBT-CSO, as well as the underlying PPO hyperparameter search space, are listed in Table~\ref{tab:pbt-hp}. At the beginning of the training, all PPO hyperparameters are generated based on logarithmic random sampling. The total number of PBT iterations is 500 for Ant, Hopper, and HalfCheetah, and 200 for Swimmer.
For experiments in Sec.~\ref{sec:speed-exp-pbt}, we record the average iteration time from 5 PBT iterations after warm-up steps. The selection ratio is 0.25, and the number of underlying workflow steps per iteration is 4 for PBT-PPO and 1024 for PBT-SAC. For PBT-PPO, the other settings are the same as above. The search space for PBT-SAC is listed in Table~\ref{tab:pbt-sac-hp} to align with the implementation in fastpbrl.

\begin{table}
    \centering
    \caption{Hyperparameters for PBT and PBT-CSO}
    \label{tab:pbt-hp}
    \resizebox{\linewidth}{!}{
        \begin{tabular}{cccccc}
            \toprule
            PBT Hyperparameter         & Value & PPO Search Space    & Value              & PPO Search Space    & Value \\ \midrule
            Population Size            & 128   & Actor Loss Weight   & [0.01,10]          & Batch Size          & 256   \\
            Warm-up Steps              & 256   & Critic Loss Weight  & [0.01,10]          & Optimizer           & Adam  \\
            Workflow Steps per Iter    & 64    & Entropy Loss Weight & [-1,-1e-5]         & Learning Rate       & 3e-4  \\
            Perturbation Factor        & 0.2   & Discount Factor     & [0.86466, 0.99999] & Gradient Norm. Clip & 10    \\
            Selection Ratio (PBT only) & 0.2   & GAE Factor          & [0.63212, 0.99999] & Timesteps per Iter  & 2048  \\
            Episodes for Meta Obj.     & 16    & Clip Epsilon        & [0.01,0.5]         & Number of Env.      & 4     \\
            Evaluation Episodes        & 128   &                     &                    & Epochs per Iter     & 4     \\ \bottomrule
        \end{tabular}
    }
\end{table}

\begin{table}
    \begin{minipage}[t]{0.45\linewidth}
        \centering
        \caption{Search Space for PBT-SAC}
        \label{tab:pbt-sac-hp}
        \begin{tabular}{cc}
            \toprule
            SAC Search Space          & Value              \\ \midrule
            Actor Loss Weight         & [0.01,10]          \\
            Critic Loss Weight        & [0.01,10]          \\
            Alpha                     & [0.00674,1]        \\
            Discount Factor           & [0.86466, 0.99999] \\
            Batch Size                & 256                \\
            Optimizer                 & Adam               \\
            Learning Rate             & 3e-4               \\
            Actor Update Interval     & 2                  \\
            Shared Replay Buffer Size & 1e6                \\ \bottomrule
        \end{tabular}
    \end{minipage}
    \hfill
    \begin{minipage}[t]{0.45\linewidth}
        \centering
        \caption{Hyperparameters for PPO}
        \label{tab:ppo-hp}
        \begin{tabular}{cccc}
            \toprule
            Hyperparameter      & Value \\ \midrule
            Actor Loss Weight   & 1.0   \\
            Critic Loss Weight  & 0.5   \\
            Entropy Loss Weight & -0.01 \\
            Discount Factor     & 0.99  \\
            GAE Factor          & 0.95  \\
            Clip Epsilon        & 0.2   \\ \bottomrule
        \end{tabular}
    \end{minipage}
\end{table}

\section{Supplementary Experimental Results}

In Sec.~\ref{sec:use-cases}, we present the performance of various EvoRL algorithms on \evorl{} framework, including ERL and PBT. This section provides supplementary experimental results that compare these EvoRL algorithms with their corresponding standalone RL algorithms. These comparisons further demonstrate the advantages of our framework in supporting EvoRL algorithms through end-to-end GPU acceleration.

\subsection{ERL Algorithms}

The ERL algorithms integrate TD3 as the underlying RL component. For comparison, we evaluate them against the original TD3 implementation under similar training settings, including identical network architectures and hyperparameters. The key differences lie in the update frequency and rollout strategies. While ERL algorithms use episodic sampling, the standalone TD3 samples two timesteps per iteration from a single environment and performs two updates (with one delayed critic update). This results in more sequential operations across the same number of episodes, thereby significantly increasing the overall training time.

\begin{figure}
    \centering
    \includegraphics[width=\linewidth]{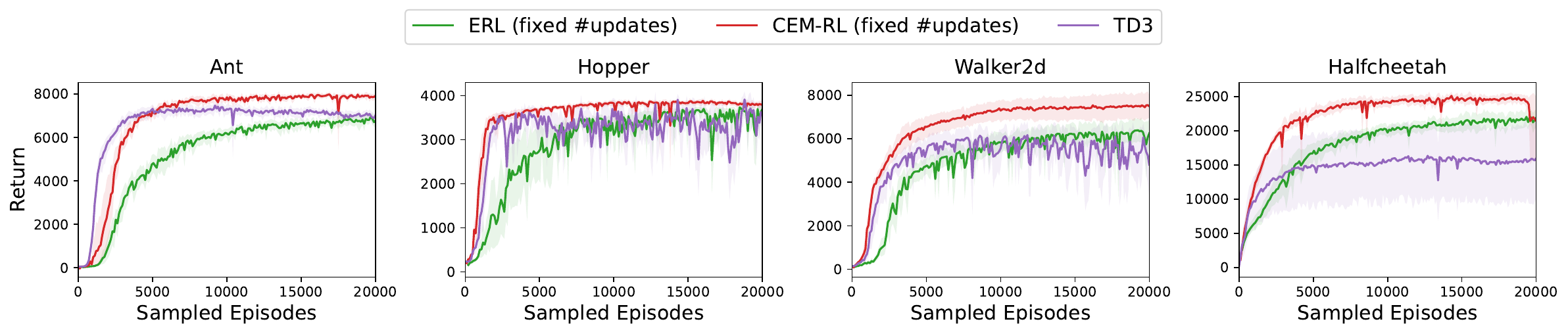}
    \caption{The benchmark of ERL and RL on robotic locomotion tasks. All ERL algorithms use TD3 as the RL component. Each algorithm is repeated with 8 different seeds. The average return and its 95\% confidence interval about the population distribution mean are reported.}
    \label{fig:erl-td3-results}
\end{figure}

\begin{figure}
    \centering
    \includegraphics[width=\linewidth]{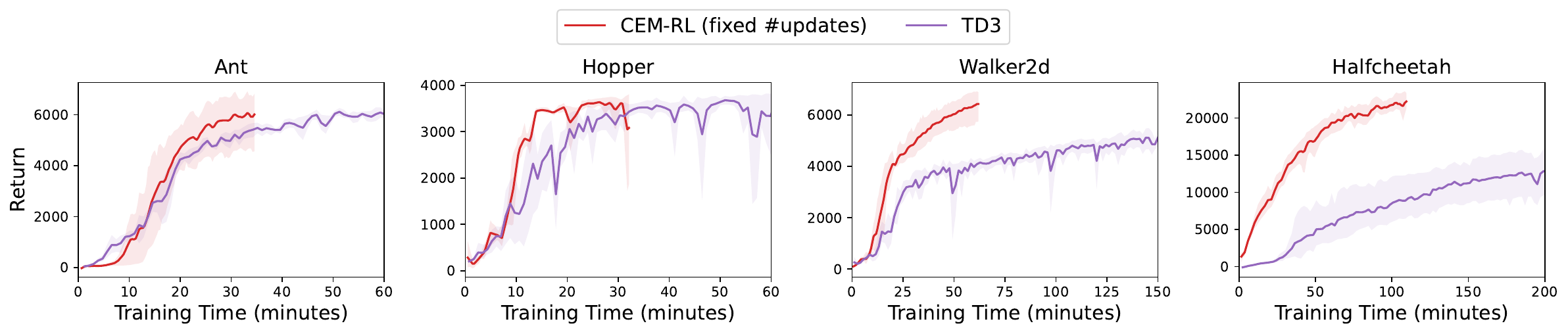}
    \caption{Evaluation of CEM-RL and TD3 on training time. Each algorithm is repeated with 4 different seeds. The average best return of the population and 95\% confidence interval are reported.}
    \label{fig:cemrl-td3-time}
\end{figure}

Due to these distinctions in data utilization strategies, it is challenging to compare performance using a uniform horizontal axis, such as the number of sampled timesteps or episodes. Therefore, we adopt alternative perspectives to evaluate performance.
Fig.~\ref{fig:erl-td3-results} shows a linear interpolation of evaluation metrics during TD3 training, plotted against the number of sampled episodes to match other ERL algorithms. Compared to TD3, CEM-RL demonstrates greater training stability and achieves higher episodic returns. Although TD3 may appear to converge at a comparable or even faster rate when assessed by the number of episodes, this can be misleading, particularly in the context of end-to-end GPU acceleration and hierarchical vectorization.
To address this, we conduct independent experiments using machines with identical hardware to evaluate on training time, as shown in Fig.~\ref{fig:cemrl-td3-time}. Both CEM-RL and TD3 implementations are built on \evorl{}, and the reported training times exclude JIT compilation and evaluation phases. The results further highlight the superiority of CEM-RL over TD3 in terms of computational efficiency.

\subsection{PBT Algorithms}

\begin{figure}
    \centering
    \includegraphics[width=\linewidth]{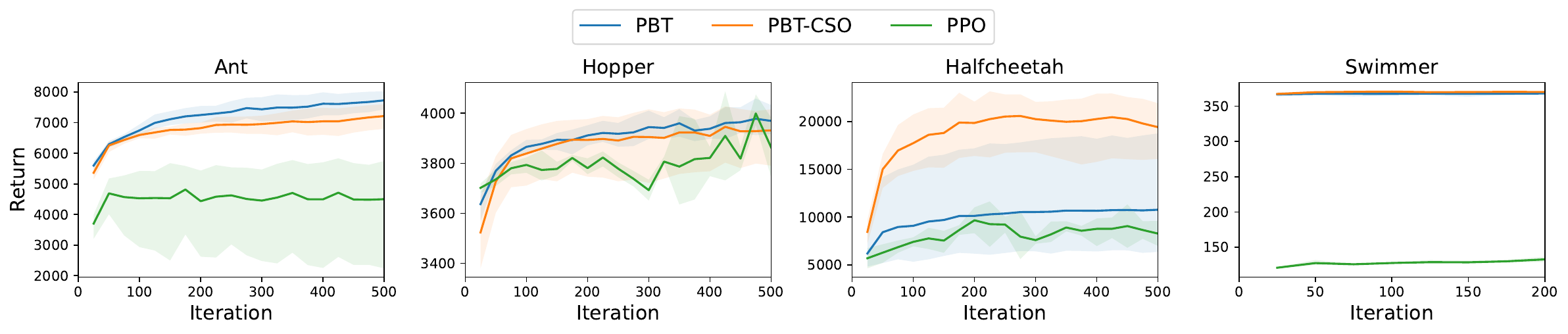}
    \caption{The benchmark of PBT, PBT-CSO, and a population of PPO with fixed hyperparameters. Each algorithm is repeated with 4 different seeds. The average best return of the population and 95\% confidence interval are reported.}
    \label{fig:pbt-results-ext}
\end{figure}

We compare the performance of PBT and PBT-CSO with the baseline PPO algorithm using fixed hyperparameters. The PPO hyperparameters, listed in Table~\ref{tab:ppo-hp}, are chosen from commonly adopted values. To ensure a fair comparison under a similar level of computational resources, we train 128 PPO agents in parallel with different random seeds. This ensemble is treated as a single training instance, and we report the highest return achieved within its population.
As shown in Fig.~\ref{fig:pbt-results-ext}, both PBT and PBT-CSO significantly outperform the baseline PPO. The improvement illustrates the effectiveness of population-based hyperparameter tuning in enhancing policy performance.

\end{document}